\documentclass{article} %
\usepackage{iclr2027_conference,times}

\usepackage{amsmath,amsfonts,bm}

\def\eqref#1{equation~\ref{#1}}

\def\1{\bm{1}}

\DeclareMathAlphabet{\mathsfit}{\encodingdefault}{\sfdefault}{m}{sl}
\SetMathAlphabet{\mathsfit}{bold}{\encodingdefault}{\sfdefault}{bx}{n}

\usepackage{hyperref}
\usepackage{url}
\usepackage{booktabs}
\usepackage{graphicx}
\usepackage{amsmath}
\usepackage{multirow}
\usepackage{algorithm}
\usepackage{algpseudocode}
\usepackage{subcaption}
\usepackage{placeins}
\usepackage{fontawesome5}
\title{Composable Decoding on the Probability Simplex: Theory and Implementation}

\author{Xiaotong Ji\thanks{Equal contribution.} \\
\textnormal{Huawei Noah’s Ark Lab}
\And
Ahmed Khaled Khamis\footnotemark[1] \\
\textnormal{\texttt{ahmedkkhamis@outlook.com}}
\And 
Rasul Tutunov \\
\textnormal{Huawei Noah’s Ark Lab}
\And
Matthieu Zimmer \\
\textnormal{Huawei Noah’s Ark Lab}
\And
Haitham Bou-Ammar \\
\textnormal{UCL Centre for AI}
}

\iclrfinalcopy %
\begin{document}

\maketitle

\begin{abstract}
Decoding for large language models is typically treated as a collection of isolated sampling strategies, with limited theoretical understanding of the behaviours they induce and how their underlying objectives relate.
We formulate decoding as an optimisation problem over next-token distributions on the probability simplex, balancing expected model score against regularisation under support constraints.
This view recovers familiar decoding methods through choices of regularisers and support constraints; more importantly, it enables new decoders to be constructed by composing distributional preferences within a single optimisation problem without external rewards, learned critics, or model parameter updates.
We introduce \href{https://github.com/KickItLikeShika/composimplex}{\textsc{CompoSimplex}~\faGithub}, a library with configurable support rules, regularisation primitives, and simplex solvers for constructing and evaluating compositional decoders.
We evaluate standard samplers, individual regularisers, and compositions across multiple models and reasoning tasks.
Our results show that compositions can realise trade-offs between single-sample quality, multi-sample quality, and diversity that are not attained by individual decoding objectives.
\end{abstract}

\section{Introduction}
Every large language model pipeline ends with a decoding step, yet decoding remains the least principled component in the stack. 
Practitioners choose from a shelf of isolated tricks: greedy decoding, temperature sampling~\citep{nadeem2020systematic}, Top-K~\citep{fan2018hierarchicalneuralstorygeneration}, Top-P (nucleus) sampling~\citep{holtzman2019curious}, and recent variants~\citep{meister2023locally,hewitt2022truncation,nguyen2025turning}, each tuned by intuition and trial-and-error.
Prior work has identified shared properties of sampling transformations and studied their quality--diversity trade-offs~\citep{nadeem2020systematic,wiher2022decoding}.
A practical challenge is to turn these insights into explicit objectives that can be configured, combined, and evaluated within a common interface.

We adopt an optimisation perspective: \emph{decoding distributions can be constructed by solving explicit optimisation problems on the probability simplex}.
The key insight is that a decoder need not choose a token directly; at each step, it can first choose a \emph{distribution} over tokens, and only then sample or take the mode.
This reframes decoding as a regularised optimisation problem: maximise expected model score subject to a regulariser that encodes structural preferences, e.g., diversity, sparsity, stability, etc.
From this single template, familiar decoding algorithms emerge as special cases: greedy decoding is the limit with no regularisation, softmax sampling is the unique optimum under negative Shannon entropy, Top-K and Top-P arise from negative entropy on restricted supports, and Sparsemax-style sparsity follows from an $\ell_2$ penalty~\citep{martins2016softmax}.
Decoders differ not by ``how they sample'' but by ``what objective they implicitly optimise''.
This formulation connects regularised prediction and optimisation-based decoding~\citep{blondel2020fenchel,noarov2025foundations,mudgal2023controlled}. Our focus is on jointly optimising complementary distributional objectives at each decoding step without external rewards, learned critics or model parameter updates.

\begin{figure}[t]
\centering
\includegraphics[trim={0em 0em 0em 0em}, clip=true, width=\linewidth]{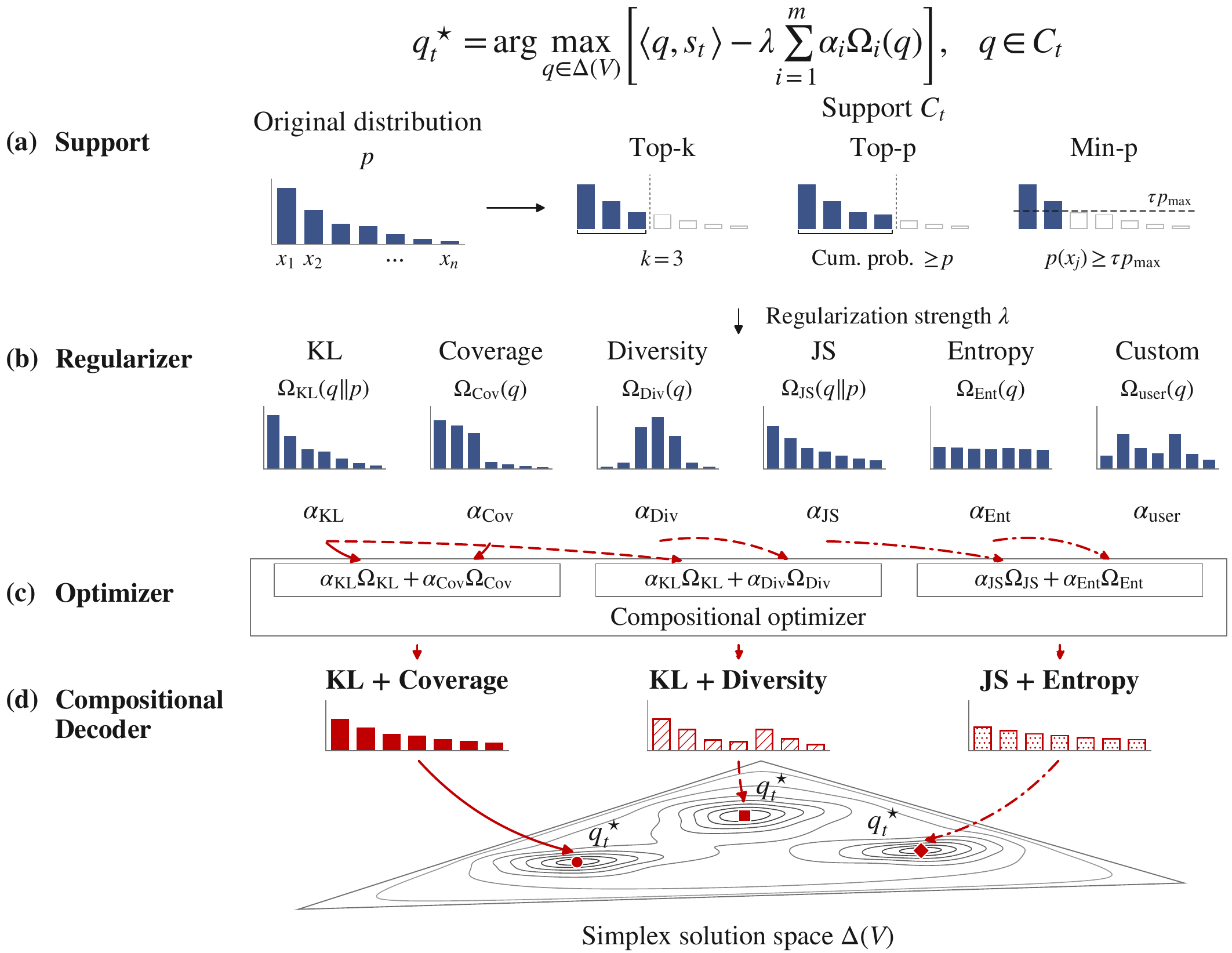}
\caption{Overview of composable decoding and the \textsc{CompoSimplex} library. Given model scores $s_t$, a decoder is configured by (a) a support constraint $C_t$, (b) regularisers $\Omega_i$ and (c) a simplex solver. The weighted regularisers are optimised jointly to construct (d) the next-token distribution $q_t^\star$.}
\label{fig:lib-structure}
\end{figure}

This optimisation view does more than unify: it provides a principled way to \emph{construct practical decoders that jointly balance multiple distributional preferences}. When a distributional preference is represented by a regulariser, multiple preferences can be \emph{composed}: a weighted sum of regularisers yields a new decoder that combines their behaviours within a single optimisation problem.
A practitioner who wants a decoder that simultaneously covers high-quality alternatives, stays anchored to the model distribution via KL divergence, and maintains entropy for diversity can declare $\Omega(q) = \alpha_1 \Omega_{\mathrm{KL}}(q) + \alpha_2 \Omega_{\mathrm{cov}}(q) + \alpha_3 \Omega_{\mathrm{ent}}(q)$ and solve on the simplex.
This compositional perspective opens up a vast design space that the community has only begun to explore.
Existing generation libraries such as Transformers~\citep{wolf2019huggingface} and vLLM~\citep{kwon2023efficient} expose sampling parameters and extensible logits processors, while \textsc{disco} provides a toolkit for distributional control~\citep{kruszewski2023disco}. 
We implement this view in \textsc{CompoSimplex}, a library with configurable support rules, distributional regularisers, and simplex solvers to examine how different distributional preferences affect the performance obtained from a language model. Figure~\ref{fig:lib-structure} illustrates how these components define and solve a composed decoding objective. 

Our contributions are as follows:
\begin{enumerate}

\item \textbf{Decoding as optimisation on the simplex.}
We formalise decoding as a regularised optimisation problem over
the probability simplex and derive the KKT optimality conditions
that recover existing decoders as special cases.

\item \textbf{Objective composition.}
We express decoder composition through a weighted sum of
regularisers that combines multiple distributional preferences
within a single optimisation problem.
The regularisers contribute additively to the optimality
conditions, and mirror ascent on the simplex provides a general
solver for composed objectives that lack closed-form solutions.
Based on this formulation, we introduce Best-of-$K$ decoding,
which combines KL regularisation with a local token-coverage
utility for a $K$-sample budget.

\item \textbf{CompoSimplex: a library for composable decoding.}
We implement the framework as a library with configurable
components, including support constraints, regularisation
primitives, and simplex solvers.
These components serve as flexible building blocks for
constructing compositional decoders through a shared
interface for Transformers and vLLM.

\item \textbf{Systematic decoding benchmark.}
We provide a decoding benchmark for evaluating model performance
across different support rules and sampling budgets, jointly
measuring accuracy, multi-sample success, and diversity.
Across four models and benchmarks, we compare standard samplers,
individual regularisers, and compositions, showing that
composition can retain the distributional preferences of
individual primitives.

\end{enumerate}

\section{Decoding on the Probability Simplex}
\label{sec:formulation}

We formulate decoding as the problem of choosing a distribution over the vocabulary at each generation step. Given a prefix $x_{<t}=(x_1,\ldots,x_{t-1})$, the language model assigns a score $s_t(v)\in\mathbb{R}$ to each
token $v$ in the vocabulary $V$ at step $t$. We view decoding as selecting a next-token distribution $q_t\in\Delta(V)$, where $\Delta (V)$ is the collection of all probability distributions defined over the vocabulary $V$.
The next token is then obtained by sampling $x_t\sim q_t$ or by selecting a mode of the distribution $x_t\in\arg\max_{v\in V} q_t(v)$. Thus deterministic and stochastic decoding differ in how the final token is selected from $q_t$, while both require the decoder to construct a distribution on the simplex.

\subsection{Decoding as Optimisation over Distributions}
\label{subsec:decoding_as_optimisation}

We define the decoding distribution as the solution of a regularised optimisation problem:
\begin{equation}
q_t^\star
=
\arg\max_{q\in\Delta(V)}
\left[
\langle q,s_t\rangle-\lambda\Omega(q)
\right],
\qquad
\text{s.t. } q\in C_t ,
\label{eq:master_problem}
\end{equation}
where $\langle q,s_t\rangle=\sum_{v\in V}q(v)s_t(v)$ is the expected model score under $q$, $\Omega(q)$ is the regulariser that encodes preferences
over the decoding distribution, and $\lambda\ge 0$ controls its strength. The set $C_t$ specifies a decoding-time feasibility constraint; for example, a support constraint restricts
sampling to a selected set of candidate tokens $S_t\subseteq V$ by
requiring $q(v)=0$ for all $v\notin S_t$. This formulation separates the model score from the decoding rule: the model provides $s_t$, while the decoder is specified by $\Omega$, $\lambda$ and $C_t$.

The score term places probability mass on high-scoring tokens, while the regulariser $\Omega(q)$ shapes how this mass is allocated across the feasible simplex. For example, negative entropy encourages probability mass to spread
across the support, whereas a divergence penalty discourages
differences from a reference distribution. In this view, a decoding rule is specified by the pair $(\Omega,C_t)$ with the regularisation strength $\lambda$, and the output of the rule is always the distribution $q_t^\star$.

\subsection{Optimality Conditions on the Simplex}
\label{subsec:optimality_conditions}
We now derive the optimality condition for Eq.~\ref{eq:master_problem}. For clarity, we first omit the support constraint $C_t$ and rewrite the maximisation as the equivalent minimisation problem
\begin{equation}
q_t^\star
=
\arg\min_{q\in\Delta(V)}
\left[
\lambda\Omega(q)-\langle q,s_t\rangle
\right].
\label{eq:min_form}
\end{equation}

Eq.~\ref{eq:min_form} can be solved as a constrained optimisation problem over the simplex. The simplex constraint consists of the normalisation condition $\sum_{v\in V}q(v)=1$ and the non-negativity conditions $q(v)\ge 0$ for all $v\in V$. We first derive the stationarity condition for coordinates in the interior of the simplex, where $q(v)>0$. On these active coordinates, the non-negativity constraints are inactive, so we can impose only the normalisation condition with a Lagrange multiplier
\begin{equation*}
\mathcal{L}(q,\eta)
=
\lambda\Omega(q)
-
\langle q,s_t\rangle
+
\eta
\left(
\sum_{v\in V}q(v)-1
\right),
\end{equation*}

where $\eta$ is the multiplier for the simplex normalisation. 
Assuming $\Omega(\cdot)$ is differentiable with respect to primal variables $q(v)$ for any coordinate with strictly positive mass ($q^{\star}_t(v) > 0$), stationarity gives
\begin{align}\label{eq:kkt_active}
    &\frac{\partial \mathcal{L}}{\partial q(v)}(q^{\star}_t)
= 0 \ \ \Longrightarrow \ \ \ s_t(v)
-
\lambda
\frac{\partial \Omega(q_t^\star)}{\partial q(v)}
=
\eta.
\end{align}
For coordinates with an optimal primal solution at the boundary $q^{\star}_t(v) = 0$, moving slightly into the feasible region must not decrease the objective, and the corresponding KKT condition gives:
\begin{align}\label{eq:kkt_inactive}
    \frac{\partial \mathcal{L}}{\partial q(v)}(q^{\star}_t) \ge 0 \ \ \Longrightarrow \ \ \ s_t(v)
-
\lambda
\frac{\partial \Omega(q_t^\star)}{\partial q(v)}
\le
\eta.
\end{align}

The quantity $s_t(v)
-
\lambda
\frac{\partial \Omega(q_t^\star)}{\partial q(v)}$ can be viewed as the regularised score of token $v$ at the optimum. 
All tokens assigned strictly positive probability have the same regularised score $\eta$, while tokens at the boundary cannot exceed this value when the derivative at zero is finite. When $C_t$ is a support constraint, the same condition applies on the feasible face of the simplex, with tokens excluded by $C_t$ fixed to zero. This optimality view recovers familiar decoding rules through specific choices of $\Omega$, $\lambda$, and $C_t$. Appendix~\ref{app:decoder_examples} provides detailed derivations for greedy, Top-K, Top-P, softmax and sparsemax decoding as special cases under our formulation. We next apply the same formulation to composed decoding objectives.

\section{Decoding by Objective Composition}
\label{sec:composition}

We use the optimisation view above to construct new compositional decoders. Many existing decoding methods are designed to control a single property, such as staying close to the base distribution, smoothing the distribution, or encouraging broader coverage across samples. Our goal is to combine such behaviours without introducing a separate decoding rule for each combination. The formulation in Eq.~\ref{eq:master_problem} makes this possible: we can express composition by combining different regularisers $\Omega$, while keeping the same score term and feasible set. In this section, we describe how composition enters the objective and its optimality condition, and how the resulting problem can be solved when no closed-form solution is available, then introduce Best-of-$K$ decoding as a special use case.

\subsection{Composition through the Regulariser}
\label{subsec:composition_regulariser}

A regulariser $\Omega$ specifies one way of shaping the decoding distribution by encoding a bias over the simplex, for example, keeping close to a reference model distribution or encouraging probability mass to cover more tokens. To obtain a decoder with multiple such characteristics, we define a composed regulariser
\begin{equation}
\Omega_\alpha(q)
=
\sum_{i=1}^{m}\alpha_i\Omega_i(q),
\label{eq:composed_regulariser}
\end{equation}
where $\Omega_i$ is the $i$-th regulariser and the weights $\alpha_i\geq 0$ satisfy $\sum_{i=1}^{m}\alpha_i=1$. 
A component may penalise an undesirable property directly, or it may be written as the negative of a quantity to be encouraged. In both cases, the composed expression is treated as a single regulariser in the original decoding objective, and we define the composed problem by substituting Eq.~\ref{eq:composed_regulariser} into Eq.~\ref{eq:master_problem}

\begin{equation}
q_t^\star
=
\arg\max_{q\in\Delta(V)}
\left[
\langle q,s_t\rangle
-
\lambda
\sum_{i=1}^{m}\alpha_i\Omega_i(q)
\right],
\qquad
\text{s.t. } q\in C_t .
\label{eq:expanded_composed_master_problem}
\end{equation}
The optimisation variable remains the distribution $q$, and the decoder still returns a distribution $q_t^\star$ on the feasible simplex. For the composed regulariser, for every active token $v$ with $q^{\star}_t(v) > 0$, the condition in Eq.~\ref{eq:kkt_active} becomes
\begin{equation}
s_t(v)
-
\lambda
\sum_{i=1}^{m}
\alpha_i
\frac{\partial \Omega_i(q_t^\star)}{\partial q(v)}
=
\eta.
\label{eq:composed_kkt_active}
\end{equation}

Feasible tokens on the boundary satisfy the corresponding inequality in Eq.~\ref{eq:kkt_inactive} when the derivatives at zero are finite. Each component regulariser contributes an additive term to the regularised score through its derivative, and the optimum balances the combined regularisation effect against the model score. This yields a simple mechanism for objective composition: multiple decoding preferences interact through additive gradient contributions within a shared optimality condition. Consequently, new decoding behaviours can be introduced by modifying or combining regularisers, without altering the underlying decoding formulation.

\subsection{Solving the Composed Objective}
\label{subsec:solving_composed_objective}
In special cases, the optimisation in Eq.~\ref{eq:min_form} can be solved analytically from the optimality condition. For example, if the derivative of $\Omega$ in Eq.~\ref{eq:kkt_active} can be inverted coordinate-wise, the normalisation constraint can determine the multiplier $\eta$ and yield a closed-form distribution. Appendix~\ref{app:decoder_examples} works through standard decoders induced by simple regularisers and support constraints. For a composed regulariser, however, Eq.~\ref{eq:composed_kkt_active} contains a sum of derivative terms, making it difficult to isolate each coordinate $q_t^\star(v)$ in closed form. We therefore solve the objective directly on the simplex.

One seemingly natural choice to tackle this problem is projected gradient ascent:
\begin{align}\label{eq:projected_gradient_objective}
    &q_{j+1}
=
\arg\max_{q\in\Delta(V)}
\left[
\left\langle \nabla f(q_j),q-q_j\right\rangle
-
\frac{1}{2\rho}
\|q-q_j\|_2^2
\right],
\end{align}
where $\rho>0$ is the step size and $f(q) =
\langle q,s_t\rangle
-
\lambda\Omega_\alpha(q)$ denotes the objective function in Eq.~\ref{eq:expanded_composed_master_problem}. This form shows that projected gradient ascent uses Euclidean distance to keep the next iterate close to $q_j$. However, the optimisation variable is a probability distribution. Euclidean distance does not reflect the geometry of the simplex, and the update requires an explicit projection step to return to a valid distribution. Mirror ascent addresses the geometry mismatch of projected gradient ascent by replacing the Euclidean distance in Eq.~\ref{eq:projected_gradient_objective} with a divergence defined on the simplex, leading to updates that remain valid distributions without an explicit Euclidean projection. 
For a strictly convex function $\psi$, define
\begin{equation}
D_\psi(q,q_j)
=
\psi(q)
-
\psi(q_j)
-
\left\langle
\nabla\psi(q_j),q-q_j
\right\rangle .
\label{eq:bregman_divergence}
\end{equation}
The mirror ascent update becomes:
\begin{equation}
q_{j+1}
=
\arg\max_{q\in\Delta(V)}
\left[
\left\langle \nabla f(q_j),q-q_j\right\rangle
-
\frac{1}{\rho}
D_\psi(q,q_j)
\right].
\label{eq:mirror_ascent_general}
\end{equation}
Using the negative entropy potential $\psi(q)=\sum_{v\in V}q(v)\log q(v)$ gives $D_\psi(q,q_j)=KL(q\|q_j)$.
Under this choice, Eq.~\ref{eq:mirror_ascent_general} reduces to the multiplicative update:
\begin{equation}
q_{j+1}
=
\frac{
q_j\odot \exp\left(\rho\nabla f(q_j)\right)
}{
\left\|
q_j\odot \exp\left(\rho\nabla f(q_j)\right)
\right\|_1
},
\label{eq:mirror_ascent_update}
\end{equation}
which preserves non-negativity and normalisation by construction. The derivation is provided in Appendix~\ref{app:mirror-ascent}. Here, $\odot$ denotes the component-wise product of two vectors in $\mathbb{R}^{|V|}$. Please note that the composed regulariser contributes to this equation via the gradient term $\nabla f(q_j)
=
s_t
-
\lambda
\sum_{i=1}^{m}\alpha_i\nabla\Omega_i(q_j).$ When feasibility conditions $C_t$ impose a support constraint, the update is applied and normalised on the feasible face of the simplex. After a fixed number of steps, the final iterate is used as the decoding distribution.

\subsection{Use Case: Best-of-\(K\) Decoding}
\label{subsec:bok}
We introduce Best-of-$K$ (BoK) decoding as an example of constructing a new decoder through objective composition in Algorithm~\ref{alg:bok}.
There are existing generation pipelines that draw multiple completions and then apply self-consistency or reranking~\citep{wang2023selfconsistencyimproveschainthought}.
In these settings, the usefulness of the candidate set depends on whether it contains good alternatives.
BoK is designed to encourage coverage across multiple samples while keeping the decoding distribution close to the model distribution. For a selected token set $S_t$ defining the support constraint $C_t$, let $p_t$ be a positive reference model distribution on $S_t$. We compose the KL regulariser with the negative of a weighted coverage utility:
\begin{equation}
\begin{aligned}
\Omega_{\mathrm{KL}}(q) = \mathrm{KL}(q\|p_t), \quad \Omega_{U_K}(q) = -\sum_{v\in S_t}w_t(v)\left[1-(1-q(v))^K\right],
\end{aligned}
\label{eq:bok-regularisers}
\end{equation}
where $w_t(v) \geq 0$. The utility adapts weighted expected coverage from classical occupancy models~\citep{boneh1997coupon}, where the bracketed term is the probability of observing token $v$ at least once in $K$ independent draws at a given prefix. Different choices of $w_t$ give the KL-Coverage and KL-Diversity variants, with the weighting schemes defined in Section~\ref{sec:library}.

Let $U_{K,t}(q)=-\Omega_{U_K}(q)$ denote the weighted coverage utility. The resulting composed objective is
\begin{equation}
q_t^\star
=
\arg\max_{q\in\Delta(S_t)}
\left[
\langle q,s_t\rangle
-
\lambda\alpha_{\mathrm{KL}}\mathrm{KL}(q\|p_t)
+
\lambda\alpha_U U_{K,t}(q)
\right]
\label{eq:bok-composed-objective}
\end{equation}

The shared mirror-ascent solver uses the gradient
\begin{equation}
\begin{aligned}
g_j(v)
={}&
s_t(v)
-
\lambda\alpha_{\mathrm{KL}}
\left(
\log\frac{q_j(v)}{p_t(v)}+1
\right)+
\lambda\alpha_U
w_t(v)K(1-q_j(v))^{K-1},
\qquad v\in S_t.
\end{aligned}
\label{eq:bok-gradient}
\end{equation}
For $K>1$, the coverage term gives diminishing returns to tokens
that are already likely to appear among the samples, while the KL term penalises departures from $p_t$. This illustrates how combining a utility function with distributional preferences can shape the next-token distribution beyond what temperature scaling alone can achieve; see Appendix~\ref{app:temperature} for a detailed discussion.

\begin{algorithm}[t]
\caption{BoK Decoder via Mirror Ascent (one decoding step)}
\label{alg:bok}
\begin{algorithmic}[1]
\Require candidate tokens $S_t$, scores $s_t$, reference $p_t$, weights $w_t$
\Require hyperparameters $K,\lambda,\alpha_{\mathrm{KL}},\alpha_U$, step size $\rho$, iterations $J$
\State Initialise $q_0 \gets p_t$
\For{$j=0,1,\ldots,J-1$}
    \For{each token $v\in S_t$}
        \State $\displaystyle g_j(v) \gets s_t(v)
        -\lambda\alpha_{\mathrm{KL}}\!\left(\log\frac{q_j(v)}{p_t(v)}+1\right)
        +\lambda\alpha_U w_t(v)K(1-q_j(v))^{K-1}$
    \EndFor
    \State $M_j \gets \max_{v\in S_t}\rho g_j(v)$ \Comment{Log-Sum-Exp stabilisation}
    \State $\widetilde q_{j+1}(v)
    \gets q_j(v)\exp\!\left(\rho g_j(v)-M_j\right),\quad v\in S_t$
    \State $q_{j+1}
    \gets \widetilde q_{j+1}/\|\widetilde q_{j+1}\|_1$
\EndFor
\State \Return $q_J$
\end{algorithmic}
\end{algorithm}

\section{CompoSimplex: A Library for Composable Decoding}
\label{sec:library}
\textsc{CompoSimplex} is an open-source library that implements the formulation in Section~\ref{sec:formulation} and objective composition in Section~\ref{sec:composition} through the configurable support rules, regularisation primitives, and simplex solvers shown in Figure~\ref{fig:lib-structure}.
A new decoder is specified by a configuration that selects its support, regularisation primitives, and optimiser settings.
The regularisation coefficient $\lambda$ controls the overall regularisation strength, and the weights $\alpha_i \geq 0$, with $\sum_i \alpha_i = 1$, control the relative contribution of each primitive.
At each generation step, these components define an optimisation problem whose solution gives the next-token distribution. The library integrates this computation with generation backends, allowing decoding methods to be constructed through configuration.

\paragraph{Support.}
A support rule selects candidate tokens $S_t\subseteq V$,
defining the constraint $C_t$ through $q(v)=0$ outside $S_t$.
We support the full vocabulary,
Top-$k$ with a fixed candidate count~\citep{fan2018hierarchicalneuralstorygeneration},
Top-$p$ based on cumulative probability mass~\citep{holtzman2019curious},
Min-$p$ with a threshold relative to the highest token
probability~\citep{nguyen2025turning},
$\eta$-sampling with an entropy-adaptive threshold~\citep{hewitt2022truncation},
and typical sampling based on proximity of token information
content to the distribution's entropy~\citep{meister2023locally}.

\paragraph{Regularisation primitives.}
An objective primitive with a computable gradient with respect to $q$ can be added and combined with others through configuration. We provide KL and JS divergences to control deviation from a reference distribution~\citep{kullback1951information,lin1991divergence}, and negative entropy to encourage broader sampling~\citep{shannon1948mathematical,jaynes1957information}.
The KL regulariser is defined in Eq.~\ref{eq:bok-regularisers}, while $\Omega_{\mathrm{JS}}(q)=\mathrm{JS}(q\|p_t)$ and $\Omega_{\mathrm{Ent}}(q)=-H(q)$.
The reference $p_t$ is the softmax of the model logits on $S_t$ with a configurable temperature. For multi-sample generation, Coverage and Diversity instantiate $\Omega_{U_K}$ in Eq.~\ref{eq:bok-regularisers} through different choices of $w_t$.
\emph{Coverage} assigns equal positive weights to the top-$r$ tokens under $p_t$ and zero elsewhere.
\emph{Diversity} uses $w_t(v)\propto d_t(v)\exp(-d_t(v)/\tau)$, where $d_t(v)$ is the gap from the largest logit and $\tau>0$, favouring alternatives with moderate logit gaps.

\paragraph{Optimiser.}
The optimiser combines the weighted gradients of the selected primitives to compute the decoding distribution. \textsc{CompoSimplex} provides closed-form solutions for supported cases, including single KL and entropy objectives, and otherwise uses mirror ascent as described in Section~\ref{subsec:solving_composed_objective}. Appendix~\ref{app:temperature} gives the KL and entropy solutions and discusses their relation to temperature scaling. For the primitives above, these updates reuse the current model logits and require no additional model forward passes. We use a small number of mirror-ascent steps to limit the added computation and report the resulting inference overhead in Section~\ref{sec:evaluation}.

\paragraph{Backend integration.}
\textsc{CompoSimplex} integrates with Hugging Face Transformers~\citep{wolf2019huggingface} and vLLM~\citep{kwon2023efficient} through custom logits processors.
The processor returns the computed log probabilities, with tokens outside the support masked, and the backend performs multinomial sampling or argmax according to the configured selection rule.
This allows the same decoder configuration to be used with either backend.

\section{Evaluation: Decoding by Objective Composition}\label{sec:evaluation}
We use \textsc{CompoSimplex} as a shared benchmarking framework to examine how different decoding objectives affect multiple dimensions of model performance. We compare standard sampling methods, individual regularisation primitives, and composed objectives in terms of accuracy, multi-sample success, and diversity, using a common evaluation setup. 
Our evaluation addresses two questions:
(i) \emph{Can composition combine the preferences encoded by individual regularisers?}
(ii) \emph{How do composed objectives affect distributional behaviour compared with individual primitives?}

\subsection{Performance Evaluation}

\paragraph{Models and benchmarks.}
We evaluate four models and benchmarks with different scales and across base models and instruct versions. We use LFM2.5-1.2B-Base~\citep{liquidai2025lfm2} on IFEval~\citep{zhou2023instruction}, which evaluates compliance with verifiable instructions; Qwen3-4B-Base~\citep{yang2025qwen3} on GPQA Diamond~\citep{rein2024gpqa}, which contains 198 science questions; and Qwen2.5-7B~\citep{qwen2025qwen25technicalreport} on MATH500~\citep{lightman2023let} for mathematical problem solving. For code generation, we evaluate the instruct model Gemma-4-26B-A4B-IT~\citep{gemmateam2026gemma4} on new problems introduced in LiveCodeBench v6~\citep{jain2024livecodebench}.

\paragraph{Decoder configurations.} The standard sampling support rules we evaluate include Top-$k$~\citep{fan2018hierarchicalneuralstorygeneration}, Top-$p$~\citep{holtzman2019curious}, Min-$p$~\citep{nguyen2025turning}, typical sampling~\citep{meister2023locally}, and $\eta$-sampling~\citep{hewitt2022truncation}. Each support rule selects the candidate tokens at a generation step.
The single primitives include KL divergence~\citep{kullback1951information}, JS divergence~\citep{lin1991divergence}, entropy~\citep{jaynes1957information}, and Coverage and Diversity primitives~\citep{boneh1997coupon}.
Composed objectives combine two or more primitives through weights $\alpha_i$. In particular, KL-Coverage and KL-Diversity instantiate the two weighted variants introduced in Section~\ref{subsec:bok}. We evaluate them alongside other compositions, compare each composition with its constituent primitives, and examine how these objectives behave across different support constraints.

\begin{figure*}[t]
  \centering
  \begin{subfigure}[t]{0.24\linewidth}
    \centering
    \includegraphics[width=\linewidth]{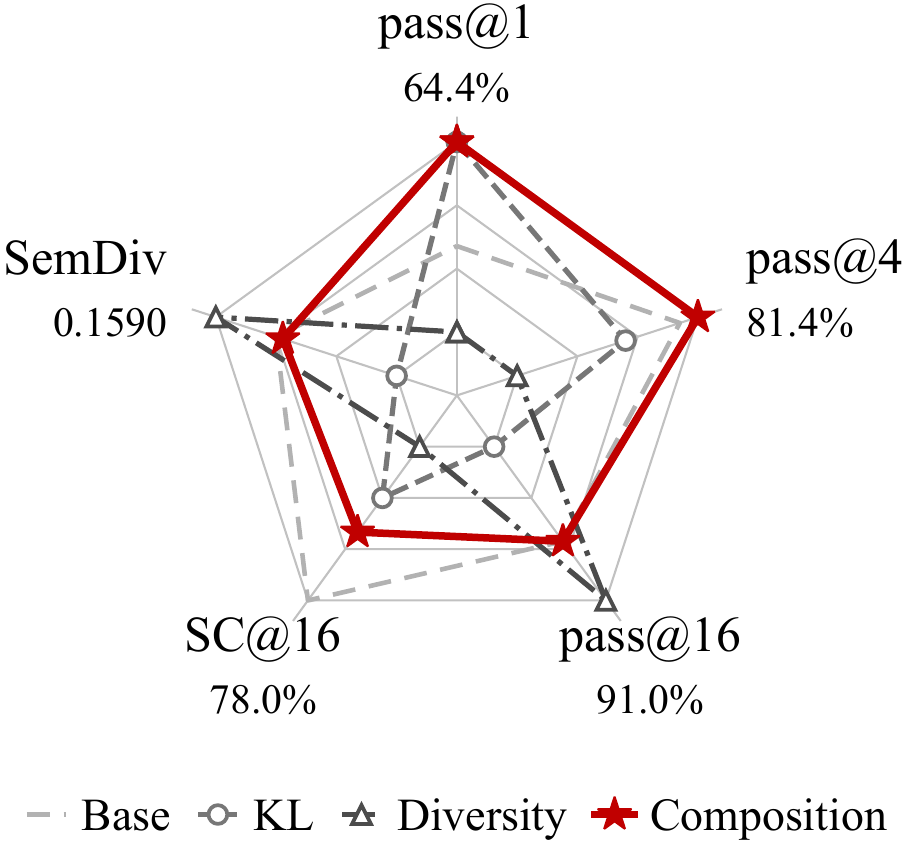}
    \caption{MATH500, Top-$k$}
    \label{fig:math-topk}
  \end{subfigure}
  \hfill
  \begin{subfigure}[t]{0.24\linewidth}
    \centering
    \includegraphics[width=\linewidth]{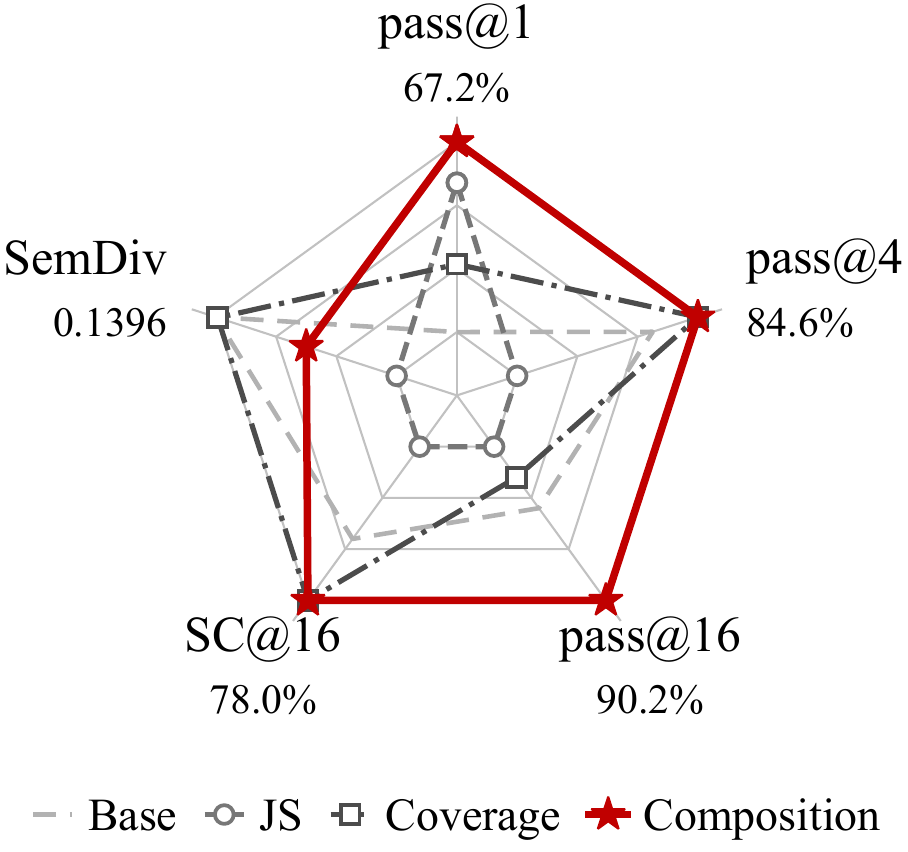}
    \caption{MATH500, Min-$p$}
    \label{fig:math-minp}
  \end{subfigure}
  \hfill
  \begin{subfigure}[t]{0.24\linewidth}
    \centering
    \includegraphics[width=\linewidth]{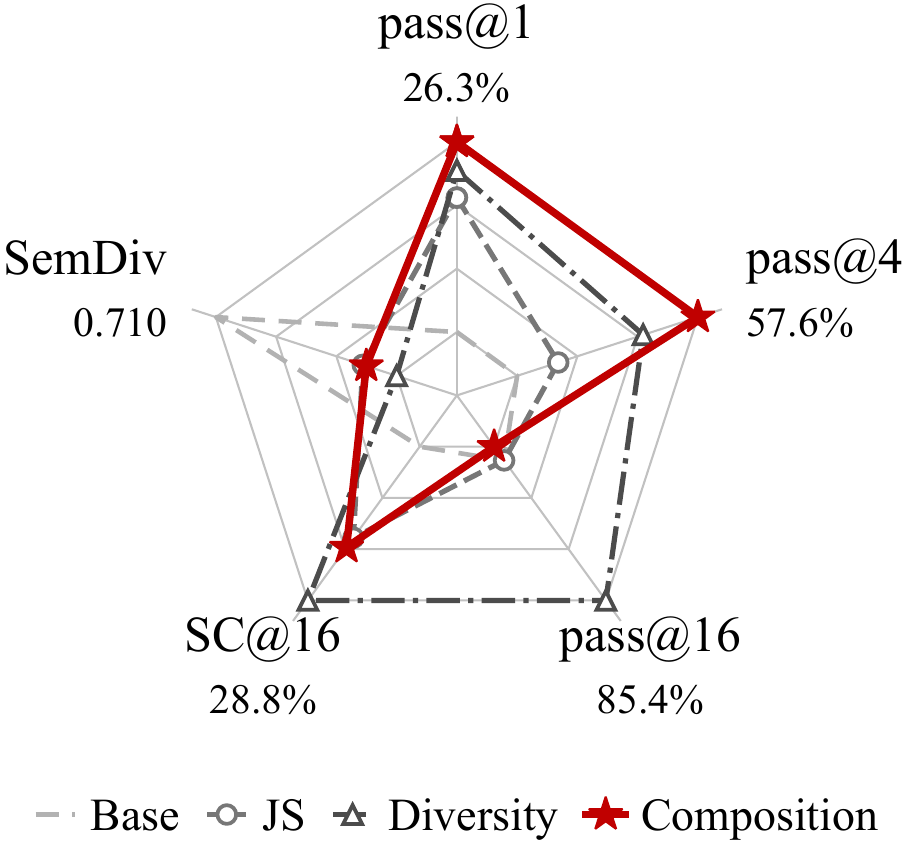}
    \caption{GPQA, Top-$k$}
    \label{fig:gpqa-topk}
  \end{subfigure}
  \hfill
  \begin{subfigure}[t]{0.24\linewidth}
    \centering
    \includegraphics[width=\linewidth]{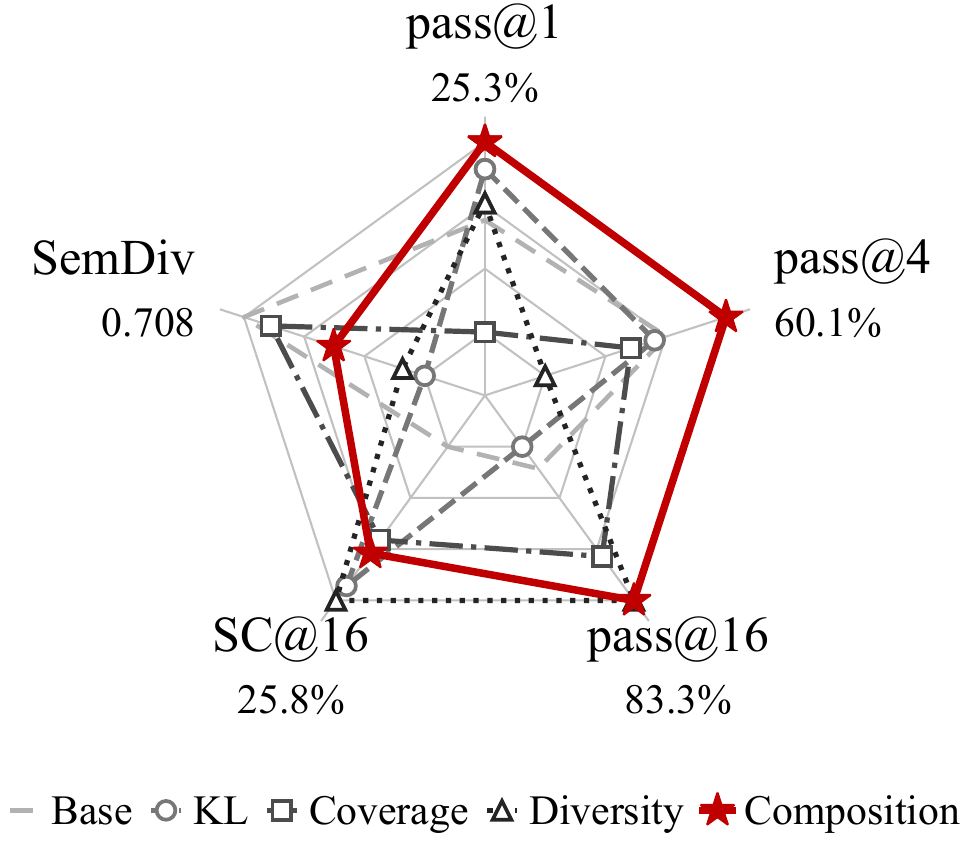}
    \caption{GPQA, Typical}
    \label{fig:gpqa-typical}
  \end{subfigure}

\begin{subfigure}[t]{0.24\linewidth}
    \centering
    \includegraphics[width=\linewidth]{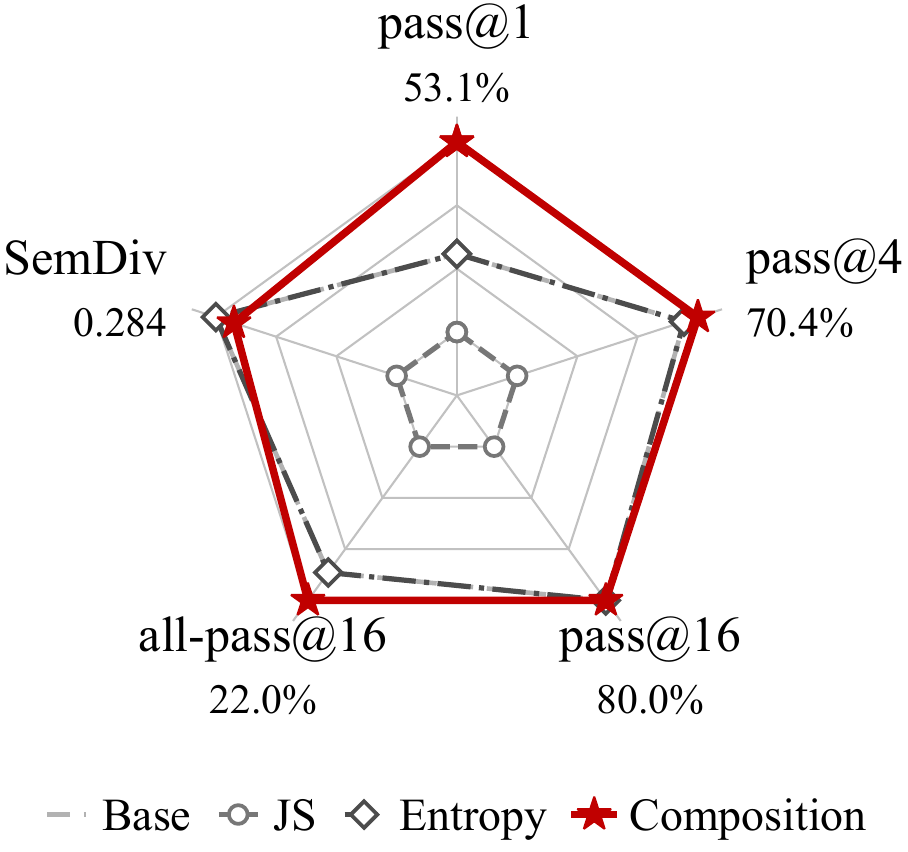}
    \caption{IFEval, Top-$p$}
    \label{fig:ifeval-topp}
    \end{subfigure}
\begin{subfigure}[t]{0.24\linewidth}
    \centering
    \includegraphics[width=\linewidth]{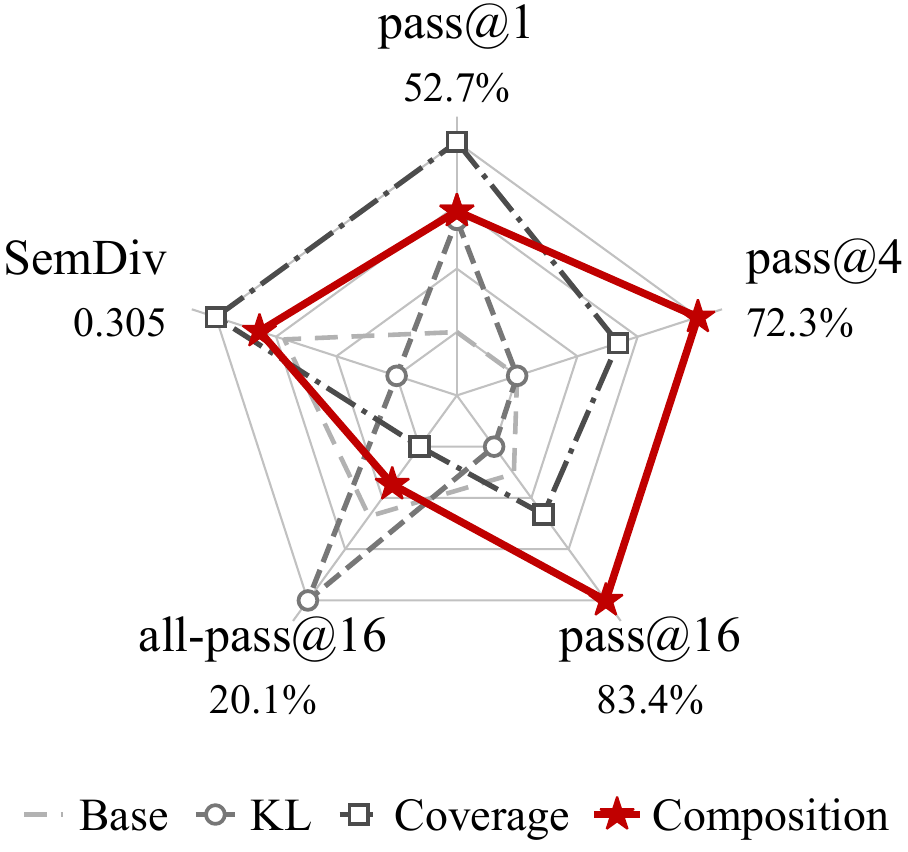}
    \caption{IFEval, $\eta$-sampling}
    \label{fig:ifeval-eta}
    
    \end{subfigure}
\begin{subfigure}[t]{0.24\linewidth}
    \centering
    \includegraphics[width=\linewidth]{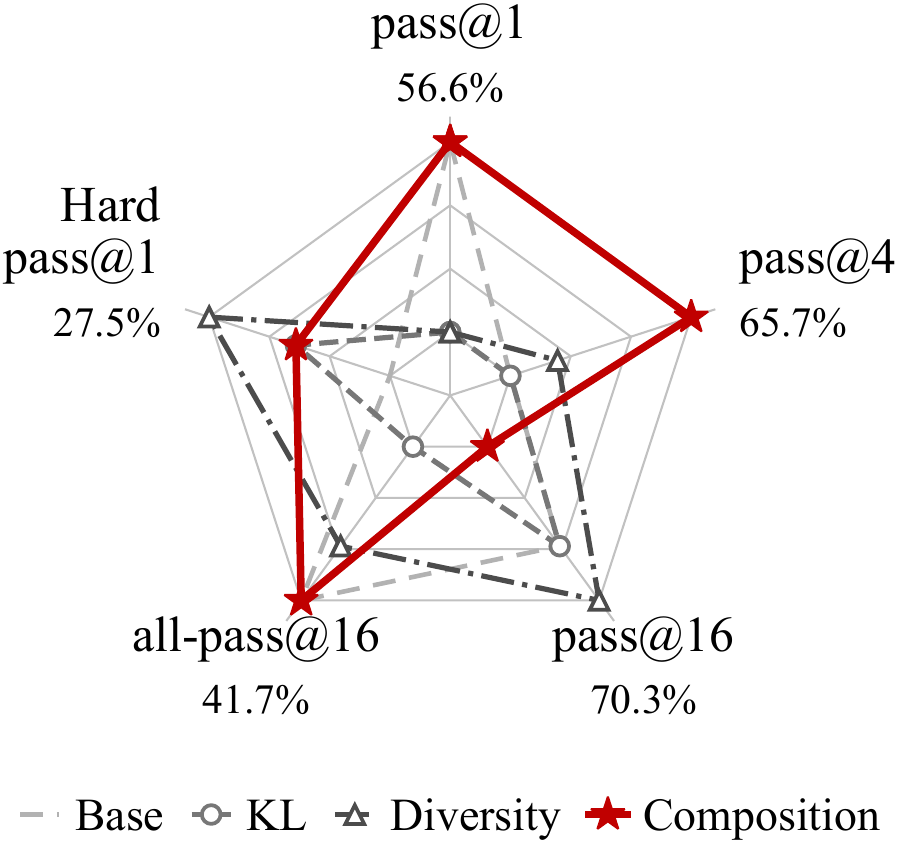}
    \captionsetup{font=scriptsize}
    \caption{LiveCodeBench, Top-$p$}
    \label{fig:livebench-topp}
    
    \end{subfigure}
\begin{subfigure}[t]{0.24\linewidth}
    \centering
    \includegraphics[width=\linewidth]{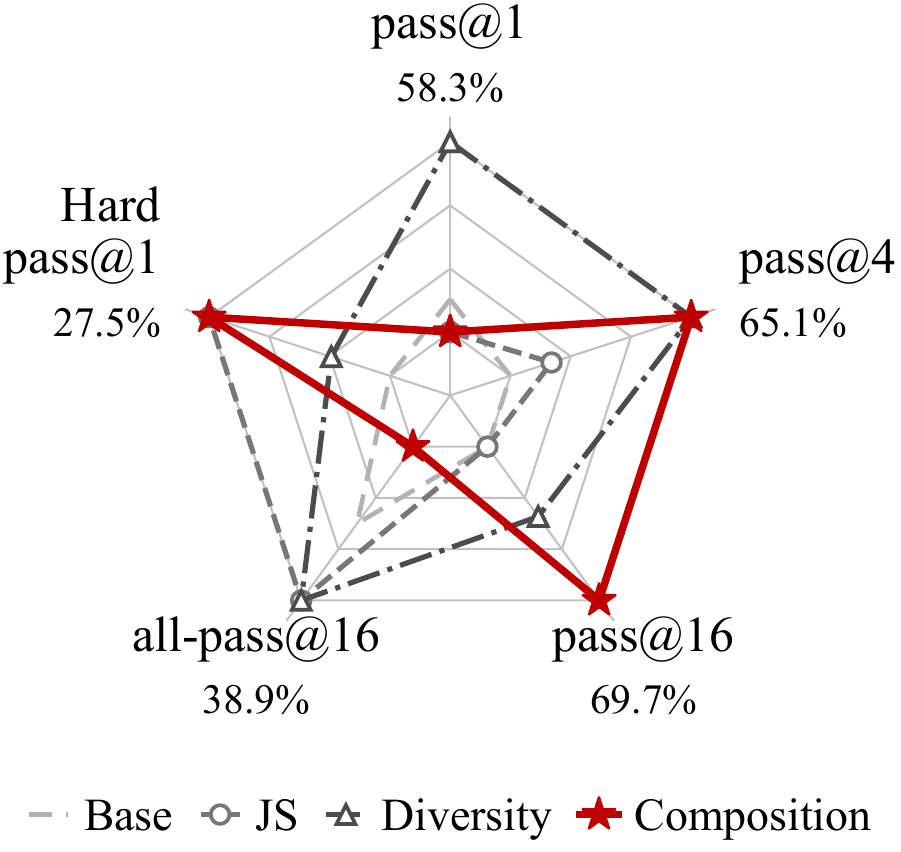}
    \captionsetup{font=scriptsize}
    \caption{LiveCodeBench, Min-$p$}
    \label{fig:livebench-minp}
    
    \end{subfigure}

\caption{Performance profiles across MATH500 (Qwen2.5-7B),
GPQA Diamond (Qwen3-4B-Base), IFEval (LFM2.5-1.2B-Base),
and LiveCodeBench v6 (Gemma-4-26B-A4B-IT).
Each panel compares a standard sampler with selected individual
and composed objectives under the indicated support rule.
The axes show the accuracy and diversity metrics labelled in each
panel.}
  \label{fig:perf_composition}
\end{figure*}

\paragraph{Evaluation metrics.}
We report pass@$k$ ($k\in\{1,4,16\}$), the fraction of prompts with at least one correct completion among the first $k$ samples. On MATH500 and GPQA Diamond, self-consistency accuracy (SC@16) uses majority voting over 16 extracted answers. Semantic diversity averages pairwise cosine distances between embeddings of sampled reasoning completions within each prompt, then across prompts. For IFEval and LiveCodeBench, all-pass@16 is the fraction of prompts whose 16 completions all pass strict prompt-level instruction checks or all test cases, respectively.
We also report LiveCodeBench's Pass@1 on hard problems.

\paragraph{Main results.}
Figure~\ref{fig:perf_composition} summarises selected performance profiles across four model--benchmark pairs and different support rules.
On MATH500 with Qwen2.5-7B and Top-$k$ support, KL+Diversity matches KL's pass@1 of $64.4\%$, $8.4$ percentage points above Diversity, while its pass@16 reaches $90.0\%$, close to Diversity's $91.0\%$ and above KL's $88.4\%$.
Across GPQA Diamond, IFEval, and LiveCodeBench, the plots also show that compositions cover a larger area than either constituent primitive in most settings, while some metrics may fall below individual regularisers.

These results show that composition can balance the preferences
of individual primitives. We also observe larger maximum gains over the corresponding base sampler in pass@1 ($+10.6$ percentage points) than in pass@16 ($+5.1$ percentage points), suggesting that regularisation can effectively concentrate probability mass on correct completions, making them easier to obtain with fewer samples.
This pattern is consistent with prior findings on decoding
and post-training~\citep{wiher2022decoding,yue2025does},
and motivates evaluating model performance across decoding
objectives and sampling budgets within a unified framework.
Full results and seed variation are reported in
Appendix~\ref{app:detailed-performance}.

\paragraph{Computational efficiency.} The generation cost for single primitives with a closed-form solution, including KL and entropy, is similar to that of the corresponding standard samplers. For single and compositional objectives without closed-form solutions, we solve the objective approximately using mirror ascent by combining the weighted gradients within each update. These updates reuse the model logits and require no additional forward passes at a given generation step. The empirical generation cost ranges from approximately 1.13$\times$ to 2.88$\times$ the corresponding base decoding method for our main evaluation. We report these costs and also examine sensitivity to the number of iterations and step size for the optimisation in Appendix~\ref{app:computational-efficiency}.

\subsection{Distributional Behaviour Analysis}\label{subsec:ablation}
We compare compositions with their constituent primitives under different regularisation strengths and composition weights. We examine how these settings change the distributional metrics and how the resulting preferences affect task performance.

\begin{figure}[h!]
\centering
\includegraphics[trim={0em 0em 0em 0em}, clip=true, width=\linewidth]{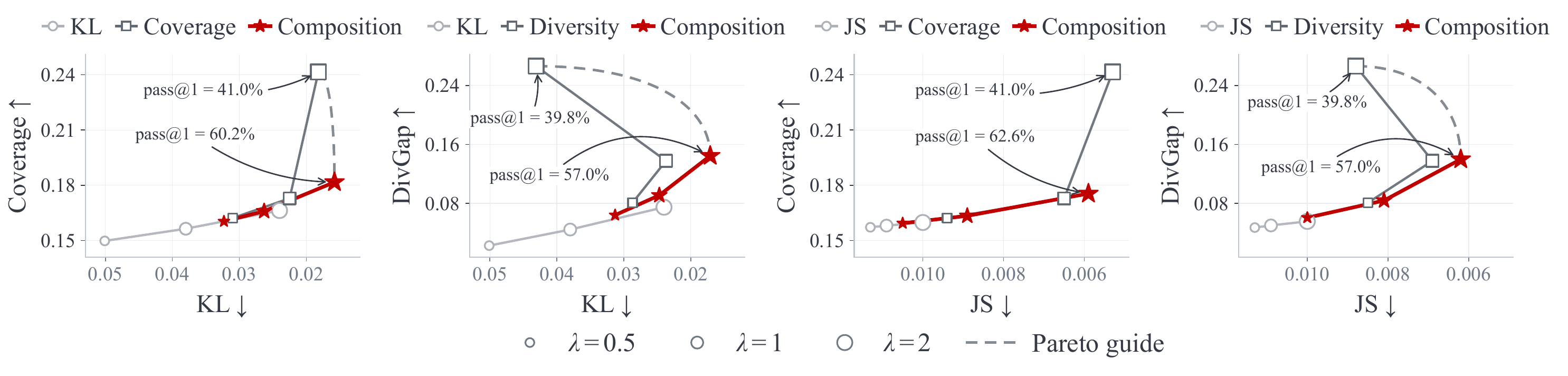}
\caption{Distributional trade-offs for individual objectives and equally weighted compositions on MATH500 with Qwen2.5-7B
and Top-$k$ support. Marker size denotes $\lambda\in\{0.5,1,2\}$; dashed curves are the Pareto guides.}
\label{fig:perf-pareto}
\end{figure}

\paragraph{Regularisation strength.}
We vary $\lambda\in\{0.5,1,2\}$ with fixed Top-$k$ support and equal composition weights on MATH500. Increasing $\lambda$ gives the selected regularisation preferences more influence relative to the model score. Across all four compositions, the corresponding utility increases while KL or JS divergence decreases, as shown in Figure~\ref{fig:perf-pareto}. At $\lambda=2$, three of the four composed points are non-dominated among the evaluated configurations in their respective divergence--utility planes.

Strong utility regularisation can nevertheless reduce task accuracy. As $\lambda$ increases from $0.5$ to $2$, Diversity's utility rises but its pass@1 falls from $62.4\%$ to $39.8\%$, while KL + Diversity retains $57.0\%$ pass@1 at $\lambda=2$. At this largest tested strength, all four compositions achieve higher pass@1 than their corresponding pure utility primitives. It is always difficult to decide the regularisation strength in the regularised objective, and the sweep also shows that compositions can retain robust task performance under strong regularisation compared with single primitives.

\paragraph{Composition weights.} We also vary the composition weights of the two BoK variants at fixed $\lambda$ and examine both distributional metrics and task performance. Across the tested weights $\alpha \in \{0, 0.25, 0.5, 1\}$, increasing the utility weight $\alpha$ monotonically raises the corresponding utility and reduces the measured KL divergence. The effects on task performance vary by metric, with no consistent improvement as $\alpha$ increases. Appendix~\ref{app:coefficient-sweeps} provides the settings and full results.

\section{Related Work}
\label{sec:related_work}

\paragraph{Sampling Methods.} Sampling methods control which tokens remain eligible and how probability mass is distributed among them.
Top-$k$ retains a fixed number of candidates~\citep{fan2018hierarchicalneuralstorygeneration}, while nucleus sampling adapts the support to retain a prescribed probability mass~\citep{holtzman2019curious}.
Temperature scaling adjusts concentration within the resulting distribution.
Studies of these transformations identify shared properties and show that their quality--diversity trade-offs depend on the task and configuration~\citep{nadeem2020systematic,wiher2022decoding}.
For multi-sample inference, \citet{du2025temperature} use an entropy-based criterion to select temperatures for answer aggregation without task-specific validation data.
These results motivate combining several distributional preferences to retain model fidelity and encourage exploration.
We express these preferences as explicit objectives and study their joint effects through objective composition.

\paragraph{Optimisation-based Decoding.}
Optimisation-based generation often targets complete sequences: DAEMON controls expected text metrics~\citep{ji2024daemon}, while power sampling sharpens the sequence distribution without external rewards~\citep{karan2026reasoning,ji2026scalablepower}.
Controlled Decoding instead applies tokenwise control using prefix value functions learned from reward supervision~\citep{mudgal2023controlled}.
Direct optimisation of the next-token distribution offers a complementary route: Bregman decoding uses a divergence and an $\ell_0$ penalty to recover a sparse distribution, with an adaptively selected support~\citep{noarov2025foundations}.
We likewise optimise a next-token distribution, but focus on jointly balancing directly computable preferences.
Their weighted combination yields a regularised simplex problem at each step, using current model scores without external rewards, learned critics, future rollouts, or model parameter updates.
Appendix~\ref{app:related_work} further compares the objectives and information used by these methods.

\section{Conclusion}
We presented a framework for decoding through regularised optimisation on the probability simplex, recovering familiar decoders as special cases and composing distributional preferences within a single objective.
Our library \textsc{CompoSimplex} implements support rules, regularisers, and solvers as flexible building blocks, with Best-of-$K$ decoding combining KL regularisation and local token coverage.
Experiments across four models and benchmarks show that composition can retain complementary strengths of individual primitives in single-sample accuracy, multi-sample success, and diversity.
Compositions can also achieve non-dominated points in distribution space while maintaining more robust task performance than single primitives under strong regularisation.
This general framework and library provide a practical basis for designing and evaluating decoding strategies through explicit, composable objectives.

\bibliography{iclr2027_conference}
\bibliographystyle{iclr2027_conference}

\clearpage
\appendix

\section{Additional Related Work}
\label{app:related_work}

\paragraph{Support Selection and Distribution Shaping.}
Desmoothing interprets truncation as removing probability mass introduced by model smoothing~\citep{hewitt2022truncation}.
Top-$n\sigma$ thresholds logits using their maximum and standard deviation, while Min-$k$ identifies truncation boundaries from relative changes in sorted logits~\citep{tang2025topnsigma,ding2026mink}.
These methods inform the choice of admissible support in our framework.
Sparsemax and $\alpha$-entmax obtain sparse distributions through regularisation, illustrating how the objective itself can determine zero-probability coordinates~\citep{martins2016softmax,peters2019sparse}.
Other theoretical accounts explain beam search through information-density objectives~\citep{meister2020if} and truncation-normalisation through approximations to a minimax strategy~\citep{chen2025decodinggame}.
Together, these connections motivate separating support selection from distribution shaping, while expressing the latter through configurable objectives.

\paragraph{Regularised Prediction and Local Decoding.}
\citet{blondel2020fenchel} define prediction as maximising a score minus an output regulariser and derive corresponding losses for supervised learning.
\citet{noarov2025foundations} apply local optimisation directly to decoding: they minimise a Bregman divergence from the model's next-token distribution together with an $\ell_0$ sparsity penalty.
Under their assumptions, the optimal support consists of the highest-probability tokens, its size can be selected adaptively, and the divergence determines how retained probabilities are reweighted.
Our focus is on a different use of local optimisation: for a chosen support, we combine divergence, entropy, and token-coverage terms to control several distributional preferences jointly.
This combination does not require external reward models, learned value functions or additional model training.

\paragraph{Sequential and Reward-Guided Decoding.}
Distributional control specifies desired output properties through constraints on expected features~\citep{khalifa2020distributional}.
DAEMON uses multiple text metrics to define a sequence-level energy-based target and approximates sampling through sampling-importance-resampling~\citep{ji2024daemon}.
COLD enforces differentiable constraints by applying Langevin dynamics to a continuous relaxation of a token sequence~\citep{qin2022cold}.
Power sampling also acts on complete-sequence distributions, but sharpens model likelihoods without external rewards or additional training, using MCMC~\citep{karan2026reasoning} or autoregressive corrections estimated from future rollouts~\citep{ji2026scalablepower}.
Some methods apply sequence-level preferences through tokenwise control: Controlled Decoding uses reward-trained prefix value functions in a KL-regularised objective and supports combinations of reward scorers~\citep{mudgal2023controlled}, while Transfer Q* estimates values for a target reward using a baseline model~\citep{chakraborty2024transferq}.
FUDGE uses learned predictors of future attributes and supports their composition~\citep{yang2021fudge}; DExperts combines expert and anti-expert language-model logits~\citep{liu2021dexperts}.
Our implemented objectives directly shape the current next-token distribution using model scores and configured references and weights, without evaluating complete trajectories or training critics.

\paragraph{Multi-sample Generation and Selection.}
Self-consistency improves answer reliability by aggregating independently sampled reasoning paths~\citep{wang2023selfconsistencyimproveschainthought}.
Stochastic beam search reduces repeated sequences through sampling without replacement~\citep{kool2019stochastic}, while arithmetic sampling coordinates draws to obtain diverse candidates~\citep{vilnis2023arithmetic}.
\citet{tang2025inference} train models to improve inference-time objectives such as pass@$k$ and majority voting.
Our question is how changing the conditional sampling distribution of a frozen model affects candidate utility at a fixed sampling budget.
The proposed BoK decoding method rewards the probability of covering weighted token alternatives in $K$ independent draws at the same prefix.
This local surrogate shapes candidate generation; its effects on accuracy and semantic diversity are tested empirically.

\paragraph{Decoding Infrastructure.}
Transformers and vLLM support custom decoding behaviour through generation settings and extensible logits processors~\citep{wolf2019huggingface,kwon2023efficient}.
\textsc{disco} makes distributional control methods accessible through reusable software components~\citep{kruszewski2023disco}.
\textsc{CompoSimplex} exposes the optimisation problem itself: users select a support, declare weighted regularisers, and choose a simplex solver.
The solver combines the regulariser gradients to optimise the declared objective jointly.
This interface connects the theoretical formulation to practical experimentation, allowing individual objectives and their compositions to be configured and compared within the same implementation.

\section{Mirror ascent closed-form expression}
\label{app:mirror-ascent}
Let us consider the mirror ascent update:
\begin{equation*}
q_{j+1}
=
\arg\max_{q\in\Delta(V)}
\left[
\left\langle \nabla f(q_j),q-q_j\right\rangle
-
\frac{1}{\rho}
D_\psi(q,q_j)
\right].
\end{equation*}
Next, we will show that using the negative entropy potential $\psi(q)=\sum_{v\in V}q(v)\log q(v)$ gives $D_\psi(q,q_j)=KL(q\|q_j)$ and the update $q_{j+1}$ allows the following closed-form expression:
\begin{equation*}
q_{j+1}
=
\frac{
q_j\odot \exp\left(\rho\nabla f(q_j)\right)
}{
\left\|
q_j\odot \exp\left(\rho\nabla f(q_j)\right)
\right\|_1
}.
\end{equation*}
The corresponding optimisation problem has the following form:
\begin{align*}
    &\min_{q(v)\ge 0, \ v\in V}\frac{1}{\rho}\sum_{v\in V}q(v)\log\frac{q(v)}{q_j(v)} - \nabla^{\mathsf{T}}f(q_j)(q - q_j) \\\nonumber
    &\text{s.t.} \ \ \sum_{v\in V}q(v) = 1.
\end{align*}
The Lagrangian has the following form:
\begin{align*}
    \mathcal{L}(q,\eta) = \frac{1}{\rho}\sum_{v\in V}q(v)\log\frac{q(v)}{q_j(v)} - \nabla^{\mathsf{T}}f(q_j)(q - q_j) + \eta\left(\sum_{v\in V}q(v) - 1\right).
\end{align*}
The first-order stationary conditions give:
\begin{align*}
    &\frac{1}{\rho}\left[\log\frac{q(v)}{q_j(v)} + 1\right] - [\nabla f(q_j)]_v + \eta = 0 \ \  \Longrightarrow \ \ q(v) = q_j(v)\exp\left(\rho[\nabla f(q_j)]_v - \rho\eta - 1\right).
\end{align*}
Using the normalisation condition $\sum_{v\in V}[q(v)] = 1$ gives:
\begin{align*}
    \sum_{v\in V}q_j(v)\exp\left(\rho[\nabla f(q_j)]_v\right)\exp(-\rho\eta -1) = 1 \ \ \Longrightarrow \ \ \exp(\rho\eta + 1) = \sum_{v\in V}q_j(v)\exp\left(\rho[\nabla f(q_j)]_v\right).
\end{align*}
This gives the final expression for the optimal primal variable:
\begin{align*}
    q(v) = \frac{q_j(v)\exp\left(\rho[\nabla f(q_j)]_v\right)}{\sum_{v\in V}q_j(v)\exp\left(\rho[\nabla f(q_j)]_v\right)}, \ \ \forall v\in V.
\end{align*}
Using non-negativity of all terms and the component-wise product $\odot$ between two vectors $q_j$ and $\nabla f(q_j)$ gives:
\begin{align*}
    q_{j+1}
=
\frac{
q_j\odot \exp\left(\rho\nabla f(q_j)\right)
}{
\left\|
q_j\odot \exp\left(\rho\nabla f(q_j)\right)
\right\|_1
}.
\end{align*}

\section{Analysis of Standard and Composed Decoders}

\subsection{Standard Decoders as Special Cases}
\label{app:decoder_examples}

The following examples recover standard decoders from
Eq.~\ref{eq:master_problem} through choices of $\Omega$, $\lambda$, and $C_t$.

\paragraph{Greedy decoding.}
Set $\Omega(q)=0$, $\lambda=0$, and $C_t=\Delta(V)$.
The objective reduces to maximising $\langle q,s_t\rangle$.
The optimality conditions become
\begin{equation}
\begin{aligned}
q_t^\star(v)>0 &\implies s_t(v)=\eta,\\
q_t^\star(v)=0 &\implies s_t(v)\leq\eta.
\end{aligned}
\end{equation}
Since at least one probability is positive,
$\eta=\max_{v\in V}s_t(v)$.
Thus any optimum places all its mass on the highest-scoring tokens.
If the maximiser $v_t^\star$ is unique, the solution is
$q_t^\star(v_t^\star)=1$ and zero elsewhere.
With tied scores, choosing a point mass on a maximiser
according to the tie-breaking rule recovers deterministic
greedy decoding.

\paragraph{Softmax sampling.}
For the negative Shannon entropy regulariser
$\Omega(q)=\sum_{v\in V}q(v)\log q(v)$, $\lambda>0$, and $C_t=\Delta(V)$,
we have $\partial\Omega(q)/\partial q(v)=1+\log q(v)$, and Eq.~\ref{eq:kkt_active} becomes
$s_t(v)-\lambda(1+\log q_t^\star(v))=\eta$.
Solving for $q_t^\star(v)$ and imposing normalisation gives
\begin{equation}
q_t^\star(v)
=
\frac{\exp(s_t(v)/\lambda)}
{\sum_{u\in V}\exp(s_t(u)/\lambda)}.
\label{eq:softmax_example}
\end{equation}
This recovers softmax sampling with temperature $\lambda$.

\paragraph{Top-K sampling.}
Let $S_t$ contain the $K$ highest-scoring tokens.
Choose negative Shannon entropy
$\Omega(q)=\sum_{v\in V}q(v)\log q(v)$, $\lambda>0$, and
\begin{equation}
C_t
=
\{q\in\Delta(V):q(v)=0\text{ for }v\notin S_t\}.
\label{eq:app_support_constraint}
\end{equation}
The objective is therefore restricted to $\Delta(S_t)$:
\begin{equation}
\max_{q\in\Delta(S_t)}
\left[
\sum_{v\in S_t}q(v)s_t(v)
-\lambda\sum_{v\in S_t}q(v)\log q(v)
\right].
\end{equation}
The entropy-regularised optimum is positive on $S_t$.
Substituting its derivative into Eq.~\ref{eq:kkt_active} gives
\begin{equation}
s_t(v)-\lambda(1+\log q_t^\star(v))=\eta
\quad\Longrightarrow\quad
q_t^\star(v)\propto\exp(s_t(v)/\lambda).
\end{equation}
Normalising over $S_t$ yields
\begin{equation}
q_t^\star(v)
=
\begin{cases}
\displaystyle
\frac{\exp(s_t(v)/\lambda)}
{\sum_{u\in S_t}\exp(s_t(u)/\lambda)},
& v\in S_t,\\[6pt]
0, & v\notin S_t.
\end{cases}
\label{eq:app_truncated_softmax}
\end{equation}
This is Top-K sampling with temperature $\lambda$;
zeros outside $S_t$ are enforced by the support constraint.

\paragraph{Top-P (nucleus) sampling.}
Top-P retains the same regulariser and changes the support
selection rule. Let $p_t^{(\lambda)}$ be the full-vocabulary
softmax distribution in Eq.~\ref{eq:softmax_example}, and order
tokens by decreasing probability.
For a threshold $p\in(0,1]$, define
\begin{equation}
m_t
=
\min\left\{
m:\sum_{i=1}^{m}p_t^{(\lambda)}(v_{(i)})\geq p
\right\},
\qquad
S_t=\{v_{(1)},\ldots,v_{(m_t)}\}.
\end{equation}
Using this $S_t$ in Eq.~\ref{eq:app_support_constraint}
gives the same restricted entropy objective as Top-K.
Its solution is therefore Eq.~\ref{eq:app_truncated_softmax},
which renormalises $p_t^{(\lambda)}$ over the nucleus.
This construction applies temperature scaling before nucleus
selection. The support is determined from the base distribution
and held fixed when optimising $q$.

\paragraph{Sparsemax.}
Choose $\Omega(q)=\frac{1}{2}\|q\|_2^2$,
$\lambda>0$, and $C_t=\Delta(V)$.
The objective becomes
\begin{equation}
\max_{q\in\Delta(V)}
\left[
\langle q,s_t\rangle-\frac{\lambda}{2}\|q\|_2^2
\right].
\end{equation}
Since $\partial\Omega(q)/\partial q(v)=q(v)$,
the two optimality conditions give
\begin{equation}
\begin{aligned}
q_t^\star(v)>0
&\implies q_t^\star(v)=\frac{s_t(v)-\eta}{\lambda},\\
q_t^\star(v)=0
&\implies s_t(v)\leq\eta.
\end{aligned}
\end{equation}
Combining them with normalisation yields
\begin{equation}
q_t^\star(v)=\frac{[s_t(v)-\eta]_+}{\lambda},
\qquad
\sum_{v\in V}[s_t(v)-\eta]_+=\lambda,
\end{equation}
where $[a]_+=\max(a,0)$ and the second equation uniquely
determines $\eta$.
Equivalently, completing the square gives
$q_t^\star=\operatorname{sparsemax}(s_t/\lambda)$,
with standard sparsemax recovered at $\lambda=1$.
Here zero probabilities arise from the quadratic regulariser
and the boundary condition, without a prescribed support set.

\subsection{Single and Composed Regularisers: Relation to Temperature Scaling}
\label{app:temperature}

We consider a single decoding step on a fixed support $S_t$, with $\lambda>0$. The scores $s_t$ are temperature-scaled model logits, and $p_t$ is the softmax of the model logits. We write $\operatorname{softmax}_{S_t}$ for normalisation over $S_t$, with zero probability outside the support.

\paragraph{KL and entropy.} KL regularisation is negative entropy regularisation with an additional linear term determined by the reference distribution.
\begin{equation}
    \mathrm{KL}(q\|p_t)=-H(q)-\langle q,\log p_t\rangle
\end{equation}

For the single-regulariser objective in Eq.~\ref{eq:master_problem}, the corresponding solutions are
\begin{equation}
    q_t^\star=
    \begin{cases}
        \operatorname{softmax}_{S_t}(s_t/\lambda),
        & \Omega(q)=-H(q),\\
        \operatorname{softmax}_{S_t}(\log p_t+s_t/\lambda),
        & \Omega(q)=\mathrm{KL}(q\|p_t).
    \end{cases}
\end{equation}

Because $\log p_t$ is a positive rescaling of $s_t$ up to an additive constant, both solutions amount to temperature scaling of the same model logits on $S_t$.
At the same $\lambda$, the extra $\log p_t$ term makes the KL solution more concentrated on high-scoring tokens.
The distinction is clearest when regularisation dominates the score term: as $\lambda\to\infty$, the entropy solution approaches the uniform distribution on $S_t$, while the KL solution approaches $p_t$. 

\paragraph{Composition goes beyond temperature scaling.} We note that including KL or entropy in a composed objective does not restrict the decoder to temperature scaling.
For BoK in Eq.~\ref{eq:bok-composed-objective}, with $\alpha_{\mathrm{KL}}>0$ and $\alpha_U>0$, the stationarity condition gives
\begin{equation}
\begin{aligned}
    q_t^\star
    &=\operatorname{softmax}_{S_t}\left(
        \log p_t+\frac{s_t}{\lambda\alpha_{\mathrm{KL}}}
        +\frac{\alpha_U}{\alpha_{\mathrm{KL}}}\nabla U_{K,t}(q_t^\star)
      \right),\\
    \big[\nabla U_{K,t}(q_t^\star)\big]_v
    &=w_t(v)K(1-q_t^\star(v))^{K-1},\qquad v\in S_t.
\end{aligned}
\end{equation}
The first two terms inside the softmax have the same temperature-scaling form as the KL decoder above.
The utility term adds a separate bonus to each token: the bonus increases with $w_t(v)$ and, for $K>1$, decreases as $q_t^\star(v)$ increases.
This gives a smaller reward for increasing a token's probability when it is already likely to appear among the $K$ samples.
Temperature scaling multiplies all score differences by the same factor.
The utility bonuses need not change these differences in the same proportion, so BoK is not restricted to temperature scaling.
The equation describes the exact optimum, which the finite-step solver approximates.
Table~\ref{tab:bok-temperature} compares the Top-$k$ baseline and BoK (KL + Diversity) across three configured temperatures.
On this grid, BoK matches or improves on the baseline with different preferences compared with the Top-$k$ decoder at each temperature.

\begin{table}[t]
    \centering
    \small
    \setlength{\tabcolsep}{7pt}
    \renewcommand{\arraystretch}{1.1}
    \caption{Top-$k$ sampling and BoK (KL + Diversity) at three
    different temperatures on MATH500 with Qwen2.5-7B.
    Both use Top-$k$ support with $k=200$; $\tau$ denotes temperature.}
    \label{tab:bok-temperature}
    \begin{tabular}{lrrrr}
        \toprule
        \textbf{Method} & \textbf{$\tau$}
        & \textbf{Pass@1 $\uparrow$}
        & \textbf{Pass@4 $\uparrow$}
        & \textbf{Pass@16 $\uparrow$} \\
        \midrule
        Top-$k$ ($k=200$)
        & 0.25 & 62.8 & 82.0 & 92.0 \\
        & 0.5 & 59.8 & 81.2 & 90.0 \\
        & 0.7 & 52.0 & 77.2 & 87.6 \\
        \midrule
        BoK (KL + Diversity)
        & 0.25 & 65.2 & 84.0 & 92.0 \\
        & 0.5 & 64.4 & 81.4 & 90.0 \\
        & 0.7 & 60.4 & 77.8 & 88.6 \\
        \bottomrule
    \end{tabular}
\end{table}

\section{Additional Experimental Results}\label{app:EvalDetails}
This section supplements the experimental results in Section~\ref{sec:evaluation}. We first describe the model checkpoints, benchmarks, decoding settings, and
evaluation protocols. We then present detailed results for selected configurations and their variation across random seeds, examine solver convergence and computational cost, and analyse the effects of regularisation strength and composition weights.

\subsection{Experimental Setup}
\label{app:experimental-setup}
Our experiments use the shared decoding and evaluation interface of \textsc{CompoSimplex}. The implementation, experiment configurations, and evaluation scripts are provided in our code repository\footnote{\url{https://github.com/KickItLikeShika/composimplex}}.
The configurations specify the model checkpoint, sampling parameters, random seed, and evaluation settings.

\paragraph{Models and benchmarks.}
We evaluate four model--benchmark pairs across instruction following, scientific reasoning, mathematical reasoning, and code generation, using base and instruction-tuned models at different scales. For instruction following, we use LFM2.5-1.2B-Base~\citep{liquidai2025lfm2}\footnote{\url{https://huggingface.co/LiquidAI/LFM2.5-1.2B-Base}} on the 541 evaluation prompts of IFEval~\citep{zhou2023instruction}\footnote{\url{https://huggingface.co/datasets/google/IFEval}}.
For scientific reasoning, we use Qwen3-4B-Base~\citep{yang2025qwen3}\footnote{\url{https://huggingface.co/Qwen/Qwen3-4B-Base}} on all 198 questions in GPQA Diamond~\citep{rein2024gpqa}\footnote{\url{https://huggingface.co/datasets/Idavidrein/gpqa}}. 
For mathematical reasoning, we use Qwen2.5-7B~\citep{qwen2025qwen25technicalreport}\footnote{\url{https://huggingface.co/Qwen/Qwen2.5-7B}} on the 500-problem MATH500 test split of MATH~\citep{hendrycks2021math}\footnote{\url{https://huggingface.co/datasets/nlile/hendrycks-MATH-benchmark}}.
For code generation, we use Gemma-4-26B-A4B-IT~\citep{gemmateam2026gemma4}\footnote{\url{https://huggingface.co/google/gemma-4-26B-A4B-it}} on the 175 new problems introduced in LiveCodeBench-v6~\citep{jain2024livecodebench}\footnote{\url{https://huggingface.co/datasets/livecodebench/code_generation_lite}}.

\paragraph{Generation and decoding settings.}
We use the Hugging Face Transformers backend for the evaluation while also providing the vLLM backend in the open-source library.
Unless otherwise stated, we sample 16 completions per prompt at temperature $T=0.5$, regularised objectives use $\lambda=1$, and compositions assign equal weights to their constituent primitives.
Top-$k$ support uses $k=200$ and Top-$p$ support uses $p=0.9$ unless otherwise indicated. 
Min-$p$ uses a relative threshold of $0.05$, typical sampling uses a cumulative mass of $0.95$, and $\eta$-sampling uses $\texttt{eta\_cutoff}=5\times10^{-4}$.
Completions terminate at a model-specific end-of-sequence token or after a maximum of 3072 completion tokens. For a fixed task, decoder comparisons use the same prompt and completion-index seed schedule. The exact benchmark prompts and model-specific chat-template settings are provided in our code repository.

We use the closed-form solution for single KL and entropy objectives. Other evaluated regularised objectives use 10 mirror-ascent updates with step size $0.1$. The solver and coefficient studies vary these settings explicitly.

\paragraph{Task-specific grading.} For MATH500, we extract the final answer with the last boxed expression. The grader normalises mathematical expressions and checks agreement with the reference through exact comparison and symbolic equivalence checks using SymPy. For GPQA, we extract an option letter from $\{A,B,C,D\}$ and compare it with the reference option. 
For MATH500, GPQA, and IFEval, we encode extracted reasoning traces using
\texttt{sentence-transformers/all-MiniLM-L6-v2}, and calculate the mean pairwise cosine distance within each prompt, averaged across prompts. For IFEval, we use the official instruction-following evaluator\footnote{\url{https://github.com/google-research/google-research/tree/master/instruction_following_eval}} and report \emph{strict prompt-level} correctness: a completion passes only if it satisfies every instruction associated with the prompt. For LiveCodeBench, we extract Python code from the generated response and use the official execution-based grader\footnote{\url{https://github.com/LiveCodeBench/LiveCodeBench}}. A completion passes only when all test cases returned by the evaluator pass; compilation errors, runtime errors, and timeouts (6s) count as failures.

\paragraph{Distributional metrics.}
At each generation step, we measure $\mathrm{KL}(q\Vert p)$, $\mathrm{JS}(q,p)$, entropy, expected coverage, and diversity-gap utility on the same selected support. For comparable coverage scores across decoders, the reported metric uses the top $k=\min(8,|S|)$ reference tokens $I_k$ and the normalisation $\mathrm{Cov}_{k}(q) =\frac{\sum_{v\in I_k}\left[1-(1-q(v))^{16}\right]}{k\left[1-(1-1/k)^{16}\right]}$. This reporting normalisation differs from the $\ell_2$-normalised weights in the Coverage objective. Diversity-gap utility uses $K=16$ and normalised weights proportional to $\Delta_v\exp(-\Delta_v)$, where $\Delta_v$ is the gap between the largest supported logit and token $v$'s logit. Each distributional metric is first averaged over generation steps within a completion and then across completions. These statistics describe the distributions encountered along each decoder's generated trajectories on average.

\subsection{Detailed Performance Evaluation}
\label{app:detailed-performance}
We provide detailed numerical results for the four model--benchmark pairs evaluated in Section~\ref{sec:evaluation}: MATH500 with Qwen2.5-7B, GPQA Diamond with Qwen3-4B-Base, IFEval with LFM2.5-1.2B-Base, and LiveCodeBench v6 with Gemma-4-26B-A4B-IT.
Each table groups the standard sampler, single primitives, and evaluated compositions within the corresponding support setting.
We compare each composition with its constituent primitives to examine which aspects of their performance profiles are retained or changed.

\paragraph{Qwen2.5-7B.}
Table~\ref{tab:math500-performance} reports single-sample and multi-sample success, self-consistency accuracy, and semantic diversity under Top-$k$ and Min-$p$ support.
Under Top-$k$, KL + Diversity retains KL's pass@1 of $64.4\%$ while increasing pass@16 from $88.4\%$ to $90.0\%$, below Diversity's $91.0\%$.
Its SemDiv also lies between the two constituents.
Under Min-$p$, JS + Coverage matches Coverage's pass@4 and SC@16 and exceeds both constituent primitives on pass@1 and pass@16, while its SemDiv lies between them.

\begin{table*}[!htbp]
\centering
\small
\setlength{\tabcolsep}{6pt}
\renewcommand{\arraystretch}{1.12}
\resizebox{0.94\linewidth}{!}{%
\begin{tabular}{lccccc}
\toprule
\textbf{Method}
& \textbf{pass@1 $\uparrow$}
& \textbf{pass@4 $\uparrow$}
& \textbf{pass@16 $\uparrow$}
& \textbf{SC@16 $\uparrow$}
& \textbf{SemDiv $\uparrow$} \\
\midrule
\multicolumn{6}{l}{\textbf{Qwen2.5-7B}} \\
\midrule
Top-k
& 59.8 & 81.2 & 90.0 & 78.0 & 0.150 \\
\midrule
\multicolumn{6}{l}{\emph{Single objective primitives}} \\
KL
& 64.4 & 80.6 & 88.4 & 76.8 & 0.132 \\
JS
& 63.6 & 82.2 & 89.0 & 76.2 & 0.136 \\
Entropy
& 59.8 & 81.2 & 90.0 & 78.0 & 0.150 \\
Coverage
& 63.4 & 82.8 & 89.4 & \textbf{78.4} & 0.157 \\
Diversity
& 56.0 & 79.4 & \textbf{91.0} & 76.2 & \textbf{0.159} \\
\midrule
\multicolumn{6}{l}{\emph{Compositions}} \\
KL + Diversity
& 64.4 & 81.4 & 90.0 & 77.2 & 0.149 \\
JS + Coverage
& 64.8 & 82.2 & 89.4 & 76.2 & 0.144 \\
JS + Coverage + Diversity
& 65.2 & 81.2 & 89.4 & 76.8 & 0.145 \\
\midrule
Min-p
& 64.4 & 84.2 & 89.6 & 77.6 & 0.1396 \\
\midrule
\multicolumn{6}{l}{\emph{Single objective primitives}} \\
KL
& 65.2 & 81.6 & 87.8 & 76.2 & 0.1275 \\
JS
& 66.6 & 83.0 & 89.2 & 77.0 & 0.1298 \\
Entropy
& 64.4 & 84.2 & 89.6 & 77.6 & 0.1396 \\
Coverage
& 65.4 & \textbf{84.6} & 89.4 & 78.0 & 0.1395 \\
Diversity
& 65.2 & 84.4 & 90.4 & 76.8 & 0.1402 \\
\midrule
\multicolumn{6}{l}{
\emph{Compositions}
} \\
JS + Coverage
& \textbf{67.2} & \textbf{84.6} & 90.2 & 78.0 & 0.1347 \\
KL + Coverage
& 64.6 & 81.6 & 89.2 & 77.0 & 0.1380 \\
KL + Diversity + Entropy
& 65.4 & 82.8 & 90.0 & 77.8 & 0.1390 \\
\bottomrule
\end{tabular}%
}
\caption{MATH500 performance of Qwen2.5-7B under Top-$k$ and
Min-$p$ support. Rows compare the standard sampler, individual
regularisers, and compositions. Accuracy values are percentages.}
\label{tab:math500-performance}
\end{table*}

\paragraph{Qwen3-4B-Base.}
Table~\ref{tab:gpqa-performance} presents the same metrics under Top-$k$ and typical support.
Under typical support, JS + Diversity retains JS's pass@1 of $26.3\%$ while increasing pass@16 from $80.8\%$ to $82.8\%$, closer to Diversity's $83.3\%$; its SC@16 lies between the constituent values.
Under Top-$k$, the same composition exceeds both constituents on pass@1 and pass@4, but records lower pass@16 than either constituent.
The resulting profile therefore depends on the support rule as well as the composed objectives.

\begin{table*}[!htbp]
\centering
\small
\setlength{\tabcolsep}{6pt}
\renewcommand{\arraystretch}{1.12}
\resizebox{0.94\linewidth}{!}{%
\begin{tabular}{lccccc}
\toprule
\textbf{Method}
& \textbf{pass@1 $\uparrow$}
& \textbf{pass@4 $\uparrow$}
& \textbf{pass@16 $\uparrow$}
& \textbf{SC@16 $\uparrow$}
& \textbf{SemDiv $\uparrow$} \\
\midrule

\multicolumn{6}{l}{\textbf{Qwen3-4B-Base}} \\
\midrule
Top-k
& 15.7
& 51.0
& 80.3
& 12.6
& \textbf{0.710} \\

\midrule
\multicolumn{6}{l}{\emph{Single objective primitives}} \\
KL
& 21.2
& 50.0
& 78.3
& 24.7
& 0.564 \\
JS
& 23.2
& 52.5
& 80.3
& 22.2
& 0.584 \\
Entropy
& 15.7
& 51.0
& 80.3
& 12.6
& \textbf{0.710} \\
Coverage
& 13.1
& 54.0
& 84.3
& 22.2
& 0.694 \\
Diversity
& 24.7
& 55.6
& \textbf{85.4}
& \textbf{28.8}
& 0.555 \\

\midrule
\multicolumn{6}{l}{\emph{Compositions}} \\
KL + Coverage
& 15.2
& 54.5
& 83.3
& 20.7
& 0.670 \\
KL + Diversity
& 21.2
& 50.5
& 83.8
& 21.2
& 0.619 \\
JS + Diversity
& \textbf{26.3}
& 57.6
& 79.8
& 23.2
& 0.581 \\
\midrule
Typical
& 20.7 & 57.6 & 80.3 & 14.1 & 0.708 \\
\midrule
\multicolumn{6}{l}{\emph{Single objective primitives}} \\
KL
& 23.7 & 57.1 & 79.8 & 24.7 & 0.561 \\
JS
& \textbf{26.3} & 54.0 & 80.8 & 23.2 & 0.578 \\
Entropy
& 20.7 & 57.6 & 80.3 & 14.1 & 0.708 \\
Coverage
& 14.1 & 56.1 & 82.3 & 21.2 & 0.686 \\
Diversity
& 21.7 & 52.5 & 83.3 & 25.8 & 0.579 \\
\midrule
\multicolumn{6}{l}{\emph{Compositions}} \\
KL + Diversity
& 24.2 & 56.1 & 79.8 & 23.2 & 0.620 \\
JS + Diversity
& \textbf{26.3} & 56.6 & 82.8 & 24.7 & 0.577 \\
KL + Coverage + Diversity
& 25.3 & \textbf{60.1} & 83.3 & 22.2 & 0.635 \\

\bottomrule
\end{tabular}%
}
\caption{GPQA Diamond performance of Qwen3-4B-Base under Top-$k$
and typical support. Rows compare the standard sampler, individual
regularisers, and compositions. Accuracy values are percentages.}
\label{tab:gpqa-performance}
\end{table*}

\paragraph{LFM2.5-1.2B-Base.}
Table~\ref{tab:ifeval-base-performance} reports strict prompt-level success and semantic diversity under Top-$p$ and $\eta$-sampling.
With $\eta$-sampling, KL + Diversity reaches pass@16 of $81.9\%$, compared with $80.0\%$ for KL and $81.5\%$ for Diversity, and records higher pass@4 and SemDiv than either constituent.
Its pass@1 and all-pass@16 are lower than those of both constituents.
Here, higher multi-sample success and semantic diversity coexist with lower reliability across repeated responses.

\begin{table*}[!htbp]
\centering
\small
\setlength{\tabcolsep}{6pt}
\renewcommand{\arraystretch}{1.12}
\resizebox{0.94\linewidth}{!}{%
\begin{tabular}{lccccc}
\toprule
\textbf{Method}
& \textbf{pass@1 $\uparrow$}
& \textbf{pass@4 $\uparrow$}
& \textbf{pass@16 $\uparrow$}
& \textbf{all-pass@16 $\uparrow$}
& \textbf{SemDiv $\uparrow$} \\
\midrule
\multicolumn{6}{l}{\textbf{LFM2.5-1.2B-Base}} \\
\midrule

Top-$p$
& 52.1
& 70.2
& 80.0
& 21.8
& 0.284 \\
\midrule
\multicolumn{6}{l}{\emph{Single objective primitives}} \\
KL
& 49.9 & 70.1 & 79.7 & 21.6 & 0.272 \\
JS
& 51.4 & 67.8 & 79.5 & 20.9 & 0.274 \\
Coverage
& 50.5 & 67.5 & 79.5 & 19.4 & 0.286 \\
Diversity
& 49.9 & 69.5 & 78.9 & 21.1 & 0.285 \\
\midrule
\multicolumn{6}{l}{\emph{Compositions}} \\
JS + Entropy
& \textbf{53.1} & 70.4 & 80.0 & \textbf{22.0} & 0.283 \\
KL + Coverage
& 53.0 & 69.7 & 77.8 & 20.5 & 0.283 \\
JS + Entropy + Diversity
& 52.7 & 70.8 & 79.5 & 21.1 & 0.282 \\

\midrule
$\eta$-sampling
& 50.5 & 68.9 & 80.6 & 17.2 & 0.294 \\
\midrule
\multicolumn{6}{l}{\emph{Single objective primitives}} \\
KL
& 51.8 & 68.9 & 80.0 & 20.1 & 0.276 \\
JS
& 51.9 & 68.9 & 80.0 & 20.9 & 0.279 \\
Coverage
& 52.7 & 70.8 & 81.5 & 14.8 & \textbf{0.305} \\
Diversity
& 51.6 & 70.8 & 81.5 & 16.6 & 0.286 \\
\midrule
\multicolumn{6}{l}{\emph{Compositions}} \\
KL + Diversity
& 50.6 & 71.5 & 81.9 & 14.8 & 0.296 \\
Coverage + Entropy
& 51.4 & 70.8 & 81.0 & 14.6 & 0.302 \\
KL + Coverage
& 51.9 & \textbf{72.3} & \textbf{83.4} & 16.1 & 0.298 \\
\bottomrule
\end{tabular}%
}
\caption{IFEval performance of LFM2.5-1.2B-Base under Top-$p$
and $\eta$-sampling support. Success uses strict prompt-level grading, and success rates
are percentages.}
\label{tab:ifeval-base-performance}
\end{table*}

\paragraph{Gemma-4-26B-A4B-IT.}
Table~\ref{tab:livecodebench-gemma-performance} reports execution-based success on the 175 problems in the LiveCodeBench v6 increment, together with all-pass@16 and pass@1 on its 80 Hard problems.
Under Top-$p$, JS + Coverage matches Coverage's pass@1 of $56.6\%$, while its Hard pass@1 lies between those of JS and Coverage.
Its pass@4 reaches $66.9\%$, above JS's $65.7\%$ and Coverage's $64.6\%$, but its pass@16 falls to $67.4\%$, compared with $69.1\%$ for both constituents.

\begin{table*}[!htbp]
\centering
\small
\setlength{\tabcolsep}{6pt}
\renewcommand{\arraystretch}{1.12}
\resizebox{0.94\linewidth}{!}{%
\begin{tabular}{lccccc}
\toprule
\textbf{Method}
& \textbf{pass@1 $\uparrow$}
& \textbf{pass@4 $\uparrow$}
& \textbf{pass@16 $\uparrow$}
& \textbf{all-pass@16 $\uparrow$}
& \textbf{Hard pass@1 $\uparrow$} \\
\midrule
\multicolumn{6}{l}{\textbf{Gemma-4-26B-A4B-IT}} \\
\midrule
Top-$p$
& 56.6 & 63.4 & 69.7 & \textbf{41.7} & 25.0 \\

\midrule
\multicolumn{6}{l}{\emph{Single objective primitives}} \\
KL
& 55.4 & 63.4 & 69.7 & 40.0 & 26.3 \\
JS
& 57.1 & 65.7 & 69.1 & 38.9 & 27.5 \\
Coverage
& 56.6 & 64.6 & 69.1 & 38.9 & 25.0 \\
Diversity
& 55.4 & 64.0 & \textbf{70.3} & 41.1 & 27.5 \\

\midrule
\multicolumn{6}{l}{\emph{Compositions}} \\
KL + Diversity
& 56.6 & 65.7 & 68.6  & \textbf{41.7} & 26.3\\
JS + Coverage
& 56.6 & \textbf{66.9} & 67.4 & 39.4 & 26.3 \\
\midrule
Min-$p$
& 55.4 & 62.9 & 68.6 & 38.3 & 23.8 \\
\midrule
\multicolumn{6}{l}{\emph{Single objective primitives}} \\
KL
& \textbf{59.4} & 65.1 & 68.6 & 40.6 & \textbf{33.8} \\
Coverage
& 57.1 & 63.4 & 69.1 & 38.3 & 26.3 \\
JS
& 54.8 & 63.4 & 68.6 & 38.9 & 27.5 \\
Diversity
& 58.3 & 65.1 & 69.1 & 38.9 & 25.0 \\
\midrule
\multicolumn{6}{l}{\emph{Compositions}} \\
KL + Coverage
& 56.0 & 65.7 & 69.1  & 38.3 & 26.3\\
JS + Diversity
& 54.8 & 65.1 & 69.7 & 37.7 & 27.5 \\
\bottomrule
\end{tabular}%
}
\caption{LiveCodeBench v6 performance of Gemma-4-26B-A4B-IT under
Top-$p$ and Min-$p$ support. Hard pass@1 is measured on the 80 Hard
problems. All reported success rates are percentages.}
\label{tab:livecodebench-gemma-performance}
\end{table*}

\paragraph{Standard deviation across random seeds.} Table~\ref{tab:seed-variance} lists standard deviations with Qwen2.5-7B on MATH500 under Top-$k$ support using seeds 0, 42 and 1234. For pass@1, pass@4, pass@16, and SC@16, most listed standard deviations are below one percentage point, with a range of $0.12$--$1.33$ percentage points.

\begin{table}[!htbp]
\centering
\small
\setlength{\tabcolsep}{4pt}
\renewcommand{\arraystretch}{1.08}
\resizebox{\linewidth}{!}{%
\begin{tabular}{lrrrrr}
\toprule
Method
& pass@1 & pass@4 & pass@16 & SC@16
& \shortstack{SemDiv\\($\times10^{-3}$)} \\
\midrule
Top-$k$

& 0.70 & 0.12 & 0.81 & 0.92 & 2.63 \\
KL

& 0.31 & 1.13 & 0.42 & 0.53 & 3.99 \\
JS

& 0.12 & 0.83 & 0.42 & 0.31 & 3.64 \\
Coverage

& 0.83 & 0.71 & 0.71 & 0.83 & 2.56 \\
Diversity

& 1.33 & 0.12 & 1.27 & 0.71 & 3.64 \\
KL + Diversity

& 0.72 & 0.71 & 1.11 & 0.31 & 3.42 \\
JS + Coverage + Diversity

& 0.90 & 0.53 & 0.42 & 0.42 & 3.19 \\
\bottomrule
\end{tabular}%
}
\caption{Sample standard deviations across seeds 0, 42, and 1234
on MATH500 with Qwen2.5-7B and Top-$k$ support. Deviations for
pass@$k$ and SC@16 are in percentage points; SemDiv deviations
are shown in units of $10^{-3}$.}
\label{tab:seed-variance}
\end{table}

\subsection{Computational Efficiency}
\label{app:computational-efficiency}

\begin{figure}[!htbp]
\centering
\includegraphics[width=\linewidth]{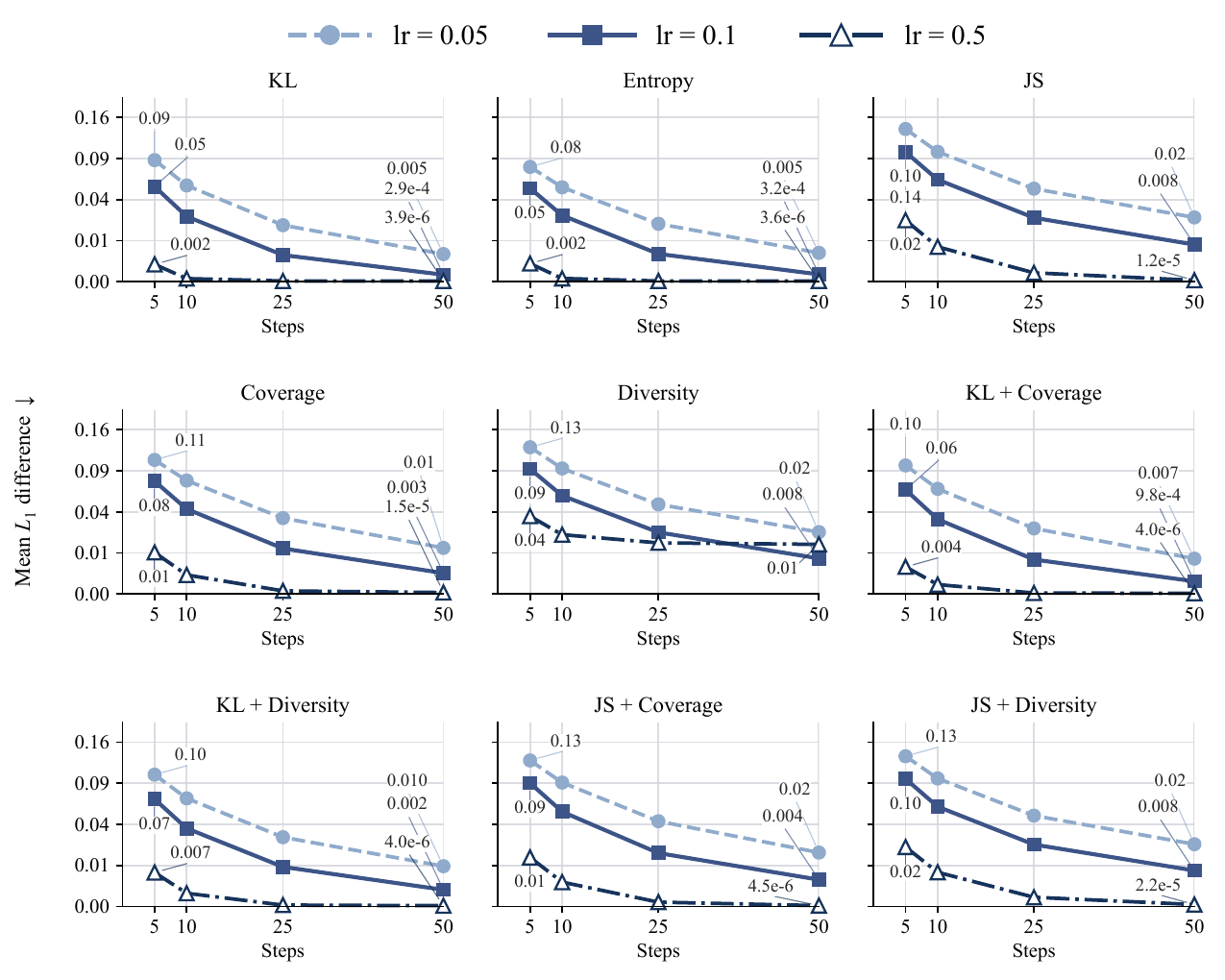}
\caption{Mirror-ascent convergence for individual and composed
objectives over 128 fixed prefixes. Each panel reports mean
$L_1$ distance to a reference optimum against the number of
updates; curves correspond to step sizes 0.05, 0.1, and 0.5.
The vertical axis uses a square-root scale with tick labels
in the original $L_1$ units.}
\label{fig:solver-convergence}
\end{figure}

\paragraph{Solver convergence.}
We examine the effect of the learning rate (step size) and iteration budget using the same 128 cached prefixes for every configuration. We fix $\lambda=1$ and vary the step size over $\{0.05,0.1,0.5\}$ and the number of updates over $\{5,10,25,50\}$. Figure~\ref{fig:solver-convergence} reports the mean $L_1$ distance between the iteratively computed token distribution and a reference optimum. We use analytic solutions for KL and Entropy and independently compute numerical reference solutions for the remaining objectives by solving the KKT conditions in float64, using bisection on the simplex normalisation multiplier and nested coordinate bisection where required.

The distance generally decreases with more updates. At step size $0.1$, the mean $L_1$ distance is below $0.009$ for every displayed objective after 50 updates. A step size of $0.5$ often reaches a smaller distance with fewer updates, but is not uniformly better. Our task-level runs use 10 updates with step size $0.1$ for iterative objectives, for which the mean distances range from $0.033$ to $0.062$. These runs therefore use finite-step approximations. KL and entropy use closed-form solutions in the task-level experiments.

\paragraph{Computational cost.}
Table~\ref{tab:compact-generation-cost} reports generation cost for selected configurations, using Top-$k$ support on MATH500 and GPQA and Top-$p$ support on IFEval and LiveCodeBench. All timing runs use seed $0$.
We compute the amortised milliseconds per output token as $1000$ divided by the recorded output token throughput.
The closed-form KL decoder has a recorded cost close to baseline decoding.
Iterative optimisation for both single and compositional objectives incurs additional cost: on MATH500, JS, Coverage and Diversity require $4.329$--$5.751$ ms/token, compared with $2.258$ ms/token for the baseline; on LiveCodeBench, they require $20.954$--$21.255$ ms/token, compared with $18.560$ ms/token.
The relative overhead varies across the recorded model and batching configurations.

\begin{table}[!htbp]
\centering
\small
\setlength{\tabcolsep}{5pt}
\renewcommand{\arraystretch}{1.08}
\begin{tabular}{lrrrr}
\toprule
Method & MATH500 & GPQA & IFEval & LiveCodeBench \\
& Top-$k$ & Top-$k$ & Top-$p$ & Top-$p$ \\
\midrule
Baseline
& 2.258 & 3.093 & 1.833 & 18.560 \\
KL
& 2.238 & 3.068 & 1.838 & 18.308 \\
JS
& 4.329 & 5.266 & 3.924 & 21.255 \\
Coverage
& 5.004 & 5.910 & 4.399 & 20.954 \\
Diversity
& 5.751 & 6.273 & 4.624 & 21.245 \\
\midrule
KL + Coverage
& 5.308 & 6.121 & 4.656 & 21.378 \\
JS + Entropy + Diversity
& 5.832 & 6.341 & 5.285 & 21.652 \\
\bottomrule
\end{tabular}
\caption{Generation cost in milliseconds per output
token for selected objectives. Columns use the model--benchmark
pair and support rule indicated in the table. The baseline uses
the base support rule without an added regulariser.}
\label{tab:compact-generation-cost}
\end{table}

\FloatBarrier

\subsection{Regularisation Strength and Composition Weights}
\label{app:coefficient-sweeps}

We study two ways of changing the decoding objective: varying the global regularisation strength $\lambda$ at fixed composition weights, and varying the relative weights at fixed $\lambda$. All experiments in this subsection use Qwen2.5-7B on MATH500, Top-$k$ support, and seed $0$. Other settings follow Appendix~\ref{app:experimental-setup}.

\paragraph{Regularisation-strength sweep.}
Table~\ref{tab:regulariser-family-sweeps} groups the results into four families: KL + Coverage, KL + Diversity, JS + Coverage, and JS + Diversity. Each block compares the two constituent primitives with their equally weighted composition at $\lambda\in\{0.5,1,2\}$. Alongside pass@1, pass@4, pass@16, SC@16, and SemDiv, the final two columns report the divergence and utility associated with that family.

Increasing $\lambda$ increases the corresponding utility and SemDiv within each of the four evaluated compositions, while reducing its KL or JS divergence.
Stronger utility regularisation does not necessarily improve accuracy: at $\lambda=2$, single Coverage and Diversity reach their highest respective utilities, but their pass@1 falls to $41.0\%$ and $39.8\%$.
The four compositions retain pass@1 between $57.0\%$ and $62.6\%$ at the same global strength, with lower utility values than the corresponding pure utility objectives.

\begin{figure}[!htbp]
\centering
\includegraphics[width=\linewidth]{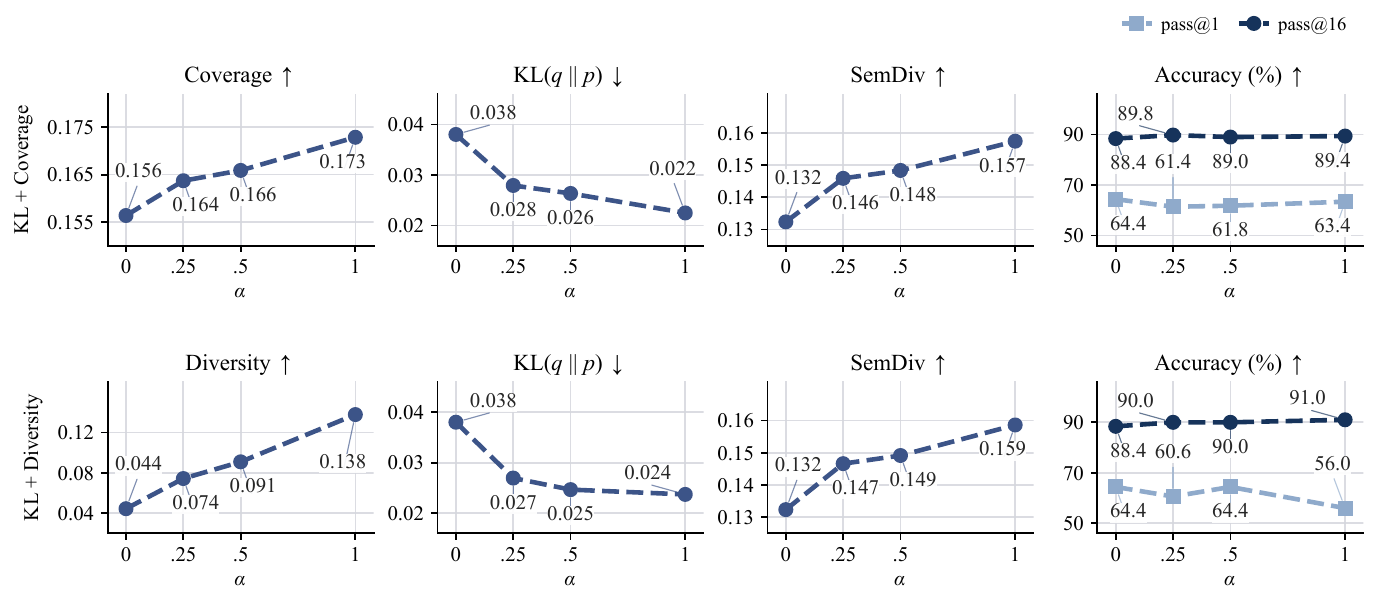}
\caption{Effect of composition weight $\alpha$ on MATH500 with
Qwen2.5-7B and Top-$k$ support at $\lambda=1$. The top row shows KL + Coverage and the bottom row KL + Diversity.
Columns show the corresponding utility, KL divergence, SemDiv,
and pass@1 and pass@16.}
\label{fig:alpha-sweep}
\end{figure}

\paragraph{Relative composition weights.} At $\lambda=1$, we vary the utility weight $\alpha\in\{0,0.25,0.5,1\}$ in KL + Coverage and KL + Diversity.
Figure~\ref{fig:alpha-sweep} reports the corresponding utility, KL divergence, SemDiv, pass@1, and pass@16.
Across the evaluated weights, increasing $\alpha$ monotonically increases the corresponding utility and decreases KL divergence in both families. These improvements show that the composed objectives shape the next-token distributions in the desired directions. These distributional improvements do not translate into consistent gains in pass@1 or pass@16 as $\alpha$ increases.
Nearby composition weights nevertheless yield broadly similar task performance, suggesting limited sensitivity to the precise choice of $\alpha$ within a small range.

\begin{table*}[!htbp]
\centering
\small
\setlength{\tabcolsep}{4.5pt}
\renewcommand{\arraystretch}{1.08}
\resizebox{0.98\linewidth}{!}{%
\begin{tabular}{lc|ccccc|cc}
\toprule

\multicolumn{9}{l}{\textbf{KL + Coverage}} \\
\cmidrule(lr){1-9}
\textbf{Method}
& $\boldsymbol{\lambda}$
& \textbf{pass@1 $\uparrow$}
& \textbf{pass@4 $\uparrow$}
& \textbf{pass@16 $\uparrow$}
& \textbf{SC@16 $\uparrow$}
& \textbf{SemDiv $\uparrow$}
& \textbf{KL $\downarrow$}
& \textbf{Coverage $\uparrow$} \\
\midrule

\multirow{3}{*}{KL}
& 0.5
& \textbf{65.8} & 82.0 & 88.0 & 73.4 & 0.111
& 0.0501 & 0.1498 \\
& 1
& 64.4 & 80.6 & 88.4 & 76.8 & 0.132
& 0.0380 & 0.1564 \\
& 2
& 59.8 & 81.2 & \textbf{90.0} & 78.0 & 0.150
& 0.0240 & 0.1663 \\

\cmidrule(lr){1-9}

\multirow{3}{*}{Coverage}
& 0.5
& 64.4 & 82.4 & 89.4 & 76.8 & 0.142
& 0.0310 & 0.1622 \\
& 1
& 63.4 & \textbf{82.8} & 89.4 & \textbf{78.4} & 0.157
& 0.0225 & 0.1729 \\
& 2
& 41.0 & 75.2 & 87.8 & 74.2 & \textbf{0.225}
& 0.0182 & \textbf{0.2415} \\

\cmidrule(lr){1-9}

\multirow{3}{*}{KL + Coverage}
& 0.5
& 62.8 & 82.2 & 88.8 & 76.0 & 0.140
& 0.0323 & 0.1604 \\
& 1
& 61.8 & 82.4 & 89.0 & 76.6 & 0.148
& 0.0263 & 0.1659 \\
& 2
& 60.2 & 81.4 & 89.2 & 77.2 & 0.168
& \textbf{0.0158} & 0.1817 \\

\midrule
\multicolumn{9}{l}{\textbf{KL + Diversity}} \\
\cmidrule(lr){1-9}
\textbf{Method}
& $\boldsymbol{\lambda}$
& \textbf{pass@1 $\uparrow$}
& \textbf{pass@4 $\uparrow$}
& \textbf{pass@16 $\uparrow$}
& \textbf{SC@16 $\uparrow$}
& \textbf{SemDiv $\uparrow$}
& \textbf{KL $\downarrow$}
& \textbf{DivGap $\uparrow$} \\
\midrule

\multirow{3}{*}{KL}
& 0.5
& \textbf{65.8} & \textbf{82.0} & 88.0 & 73.4 & 0.111
& 0.0501 & 0.0226 \\
& 1
& 64.4 & 80.6 & 88.4 & 76.8 & 0.132
& 0.0380 & 0.0444 \\
& 2
& 59.8 & 81.2 & 90.0 & \textbf{78.0} & 0.150
& 0.0240 & 0.0747 \\

\cmidrule(lr){1-9}

\multirow{3}{*}{Diversity}
& 0.5
& 62.4 & 81.4 & 89.2 & 76.8 & 0.144
& 0.0287 & 0.0808 \\
& 1
& 56.0 & 79.4 & \textbf{91.0} & 76.2 & 0.159
& 0.0237 & 0.1377 \\
& 2
& 39.8 & 71.2 & 86.8 & 71.2 & \textbf{0.185}
& 0.0431 & \textbf{0.2662} \\

\cmidrule(lr){1-9}

\multirow{3}{*}{KL + Diversity}
& 0.5
& 64.2 & 81.4 & 89.2 & 77.4 & 0.141
& 0.0313 & 0.0640 \\
& 1
& 64.4 & 81.4 & 90.0 & 77.2 & 0.149
& 0.0247 & 0.0909 \\
& 2
& 57.0 & 79.4 & 89.6 & 77.0 & 0.166
& \textbf{0.0171} & 0.1445 \\
\midrule
\multicolumn{9}{l}{\textbf{JS + Coverage}} \\
\cmidrule(lr){1-9}
\textbf{Method}
& $\boldsymbol{\lambda}$
& \textbf{pass@1 $\uparrow$}
& \textbf{pass@4 $\uparrow$}
& \textbf{pass@16 $\uparrow$}
& \textbf{SC@16 $\uparrow$}
& \textbf{SemDiv $\uparrow$}
& \textbf{JS $\downarrow$}
& \textbf{Coverage $\uparrow$} \\
\midrule

\multirow{3}{*}{JS}
& 0.5
& 63.6 & 81.2 & 88.6 & 76.8 & 0.134
& 0.0113 & 0.1572 \\
& 1
& 63.6 & 82.2 & 89.0 & 76.2 & 0.136
& 0.0109 & 0.1581 \\
& 2
& 60.6 & 82.0 & 89.2 & 76.6 & 0.140
& 0.0100 & 0.1598 \\

\cmidrule(lr){1-9}

\multirow{3}{*}{Coverage}
& 0.5
& 64.4 & 82.4 & 89.4 & 76.8 & 0.142
& 0.0094 & 0.1622 \\
& 1
& 63.4 & \textbf{82.8} & 89.4 & \textbf{78.4} & 0.157
& 0.0065 & 0.1729 \\
& 2
& 41.0 & 75.2 & 87.8 & 74.2 & \textbf{0.225}
& \textbf{0.0053} & \textbf{0.2415} \\

\cmidrule(lr){1-9}

\multirow{3}{*}{JS + Coverage}
& 0.5
& 63.8 & 82.0 & \textbf{90.2} & 75.8 & 0.138
& 0.0105 & 0.1594 \\
& 1
& \textbf{64.8} & 82.2 & 89.4 & 76.2 & 0.144
& 0.0089 & 0.1634 \\
& 2
& 62.6 & 81.0 & 89.8 & 75.8 & 0.162
& 0.0059 & 0.1755 \\
\midrule
\multicolumn{9}{l}{\textbf{JS + Diversity}} \\
\cmidrule(lr){1-9}
\textbf{Method}
& $\boldsymbol{\lambda}$
& \textbf{pass@1 $\uparrow$}
& \textbf{pass@4 $\uparrow$}
& \textbf{pass@16 $\uparrow$}
& \textbf{SC@16 $\uparrow$}
& \textbf{SemDiv $\uparrow$}
& \textbf{JS $\downarrow$}
& \textbf{DivGap $\uparrow$} \\
\midrule

\multirow{3}{*}{JS}
& 0.5 & \textbf{63.6} & 81.2 & 88.6 & 76.8 & 0.134
& 0.0113 & 0.0471 \\
& 1 & \textbf{63.6} & 82.2 & 89.0 & 76.2 & 0.136
& 0.0109 & 0.0500 \\
& 2 & 60.6 & 82.0 & 89.2 & 76.6 & 0.140
& 0.0100 & 0.0555 \\

\cmidrule(lr){1-9}

\multirow{3}{*}{Diversity}
& 0.5 & 62.4 & 81.4 & 89.2 & 76.8 & 0.144
& 0.0085 & 0.0808 \\
& 1 & 56.0 & 79.4 & \textbf{91.0} & 76.2 & 0.159
& 0.0069 & 0.1377 \\
& 2 & 39.8 & 71.2 & 86.8 & 71.2 & \textbf{0.185}
& 0.0088 & \textbf{0.2662} \\

\cmidrule(lr){1-9}

\multirow{3}{*}{JS + Diversity}
& 0.5 & 62.8 & \textbf{82.4} & 89.8 & 76.4 & 0.139
& 0.0100 & 0.0605 \\
& 1 & 63.0 & \textbf{82.4} & 89.0 & \textbf{77.0} & 0.145
& 0.0081 & 0.0838 \\
& 2 & 57.0 & 79.8 & 89.4 & 76.4 & 0.160
& \textbf{0.0062} & 0.1398 \\
\bottomrule
\end{tabular}%
}
\caption{Effect of regularisation strength
$\lambda\in\{0.5,1,2\}$ on KL--Coverage, KL--Diversity,
JS--Coverage, and JS--Diversity on MATH500 with Qwen2.5-7B
and Top-$k$ support. Each family reports task performance,
semantic diversity, the indicated divergence, and its
coverage or diversity utility.}
\label{tab:regulariser-family-sweeps}
\end{table*}

\end{document}